\documentclass[11pt, a4paper, logo, copyright, nonumbering]{xiaomi}

\usepackage[authoryear, sort&compress, round]{natbib}
\usepackage{dblfloatfix}
\usepackage[normalem]{ulem}
\usepackage{caption}
\usepackage{dramatist}
\usepackage{xspace}
\usepackage{pifont}
\usepackage{multirow}
\usepackage{tcolorbox}
\usepackage{xltabular}
\usepackage{longtable}
\usepackage{hyperref}
\usepackage{wrapfig}
\usepackage{enumitem}
\usepackage{amsfonts}
\usepackage{amsmath}
\usepackage{amssymb}
\usepackage{lineno}
\usepackage{multirow}
\usepackage{adjustbox}
\usepackage{graphicx}

\usepackage{CJKutf8}
\usepackage{setspace}
\usepackage{makecell}
\usepackage{subcaption}
\usepackage{multicol}
\usepackage{xcolor}
\usepackage{makecell}

\usepackage{siunitx}

\usepackage{hyperref}

\usepackage{orcidlink}

\usepackage{adjustbox}
\usepackage{threeparttable}
\usepackage[table]{xcolor}
\usepackage{multirow}
\usepackage{booktabs}
\usepackage{array}
\usepackage{caption}
\usepackage{float}
\usepackage[ruled,vlined]{algorithm2e}
\usepackage{wrapfig}
\usepackage{makecell}
\usepackage{tabularx}
\usepackage{algorithmic}

\usepackage[utf8]{inputenc}
\usepackage[T1]{fontenc}
\usepackage{hyperref}
\usepackage{cleveref}
\usepackage{url}
\usepackage{booktabs}
\usepackage{amsfonts}
\usepackage{nicefrac}
\usepackage{xcolor}
\usepackage{colortbl}
\usepackage{tcolorbox}
\definecolor{BrickRed}{rgb}{.72,0,0}
\definecolor{darkgreen}{rgb}{0.0, 0.5, 0.0}
\definecolor{ForestGreen}{RGB}{34,139,34}
\definecolor{LakeBlue}{RGB}{0,61,153}
\definecolor{MiOrange}{RGB}{255,225,204}
\definecolor{Hex}{RGB}{225,213,231}
\definecolor{darkgreen}{RGB}{0,100,0}

\hypersetup{
    colorlinks=true,
    allcolors=LakeBlue
}

\newcommand{\modelname}{DeltaV}

\title{\centering \modelname{}: Thinking with Visual State Updates in Unified Large Multimodal Models}

\author{
Pengjie Wang$^{1,*,\ddagger}$,
Linger Deng$^{1,*,\ddagger}$,
Zujia Zhang$^{1}$,
Shaojie Zhang$^{2}$,
Zhenbo Luo$^{2}$,
Pei Fu$^{2}$,
Jian~Luan$^{2}$,
Xiang Bai$^{1}$,
Yuliang Liu$^{1,\dagger}$
\par
{\normalfont\sffamily\fontsize{11}{15}\selectfont
$^{1}$Huazhong University of Science and Technology
\quad
$^{2}$MiLM Plus, Xiaomi Inc.
}
}

\usepackage{microtype}
\begin{document}

\begin{abstract}

    Current Unified Large Multimodal Models (ULMMs) support interleaved multimodal reasoning through textual reasoning and intermediate visual states, but typically generate each visual state as a full image.
    This full-image generation paradigm introduces substantial visual-token redundancy and dilutes supervision on sparse yet reasoning-critical state transitions.
    We propose \modelname{}, a ULMM that replaces full-image generation with visual updates. Conditioned on historical visual states, \modelname{} incrementally predicts compact update tokens that capture the visual changes across reasoning steps, avoiding repeated modeling of unchanged content. To align the token budget of each update with the magnitude of visual change, \modelname{} introduces a temporal similarity (TSIM) Router, which stops allocating tokens once the marginal reconstruction gain falls below a threshold. To support more diverse and generalizable reasoning, we further construct StructCoT, a large-scale interleaved multimodal reasoning dataset with 1.05M samples spanning 44 task domains. Experiments show that the visual-update paradigm reduces newly generated visual tokens by 55.6\% on average without compromising reconstruction fidelity, and improves multimodal reasoning by 3.3\% over full-image generation. Trained with StructCoT and large-scale multimodal data, \modelname{}-2B further outperforms substantially larger open-source models by 8.4\% on in-domain multimodal reasoning evaluations and surpasses the comparable-scale Qwen3-VL-2B by 5.9\% on external multimodal reasoning and understanding benchmarks. Code, models, and StructCoT will be released at \url{https://github.com/Pengjie-W/DeltaV}.

\end{abstract}

\maketitle

\renewcommand{\thefootnote}{\fnsymbol{footnote}}
\footnotetext[1]{Equal contribution.}
\footnotetext[3]{Work done during internships at Xiaomi Inc.}
\renewcommand{\thefootnote}{\arabic{footnote}}

\section{Introduction}

Multimodal Large Language Models (MLLMs)~\citep{wang2025internvl3, wu2024deepseek, Qwen3-VL, li2024monkey, deng2026geofocus} have demonstrated strong performance on general vision-language benchmarks~\citep{liu2024mmbench, chen2024we}.
However, they still struggle with complex tasks,
such as embodied intelligence, scientific problem solving, and spatial intelligence, that require multi-step reasoning over evolving visual structures~\citep{hao2025can, jiang2025mme}.
To address this limitation, prior studies introduce intermediate visual states into the reasoning process, forming a think-and-sketch paradigm known as interleaved multimodal reasoning~\citep{zheng2025deepeyes, li2025imagine}.

\begin{figure*}[t!]
    \centering
    \includegraphics[width=1\linewidth]{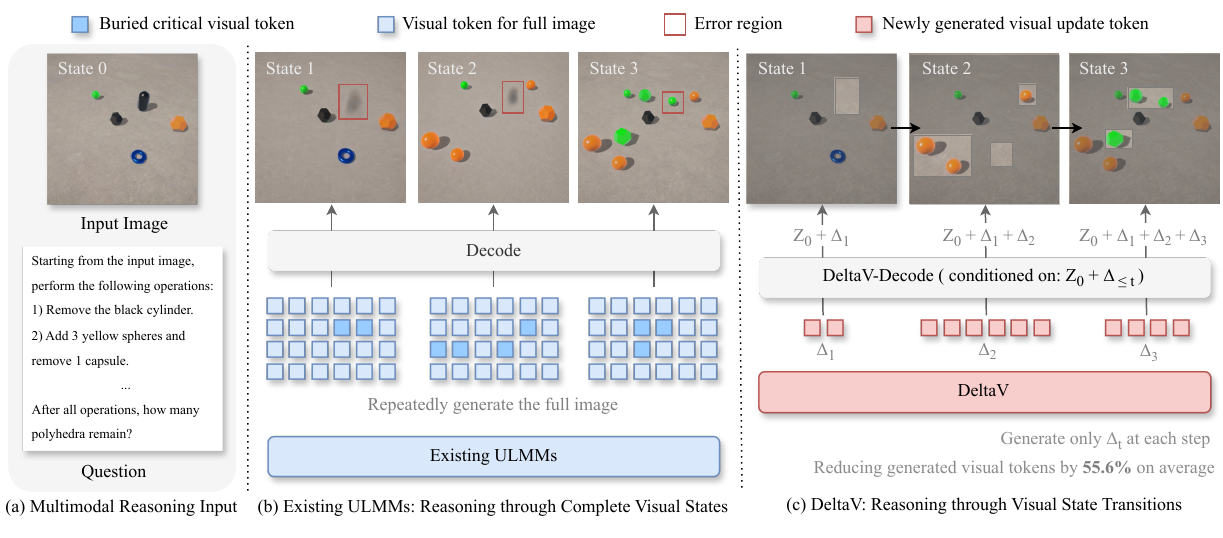}
    \caption{From generating complete visual states to modeling how visual states evolve. Comparison between existing ULMMs and \modelname{} for interleaved multimodal reasoning. Given the initial visual state tokens $Z_0$ tokenized from the input image, existing ULMMs generate each intermediate visual state as a full image, whereas \modelname{} thinks with variation-aware visual updates and dynamically allocates tokens to critical visual changes.}
    \label{fig:intro}
\end{figure*}

Unified Large Multimodal Models (ULMMs) provide a promising end-to-end framework for interleaved multimodal reasoning~\citep{chen2025janus,deng2025emerging}, as they unify language reasoning and visual generation within a single model.
While their native visual generation capability is impressive, its potential for supporting reasoning remains underexplored~\citep{zou2025uni}.
Even after fine-tuning on multimodal reasoning tasks~\citep{li2025zebra,gu2025thinkmorph,li2025imagine},
their performance may remain constrained by underlying visual generation paradigms.
As shown in Fig.~\ref{fig:intro}(b), existing ULMMs typically require hundreds of visual tokens or multiple denoising steps to produce each intermediate image, imposing
a
substantially higher cost than textual reasoning steps~\citep{zhao2025unified}.
Intuitively, costly pixel-level generation should not dominate tasks
centered on semantic reasoning.
This has given rise to the view that current pixel-level generation paradigms do not effectively facilitate multimodal reasoning~\citep{zhang2026thinkjepa,tong2026beyond}.

We argue that this bottleneck
stems not from pixel-level generation itself, but from treating each visual reasoning step as full-image generation.
In interleaved settings, consecutive visual states are often highly correlated, and reasoning is driven primarily by sparse but semantically critical changes.
Full-image generation does not fully exploit this continuity, repeatedly modeling mostly unchanged or task-irrelevant content.
This not only incurs substantial computational overhead but also dilutes the supervision signal on critical visual transitions, biasing optimization toward low-level reconstruction rather than reasoning-relevant state updates.

From a state-transition perspective, interleaved visual reasoning is better understood as the evolution of a visual state through sparse, task-relevant changes, rather than as a sequence of independently generated images. Each intermediate visual state contains both persistent content inherited from previous steps and newly introduced evidence required by the current reasoning step. Reconstructing the entire state at every step therefore spends much of the learning signal on already established content and makes it difficult to assign credit to the causal visual transitions that actually drive reasoning. Modeling visual updates instead concentrates supervision on the changes that transform one reasoning state into the next, yielding a denser and more reasoning-relevant training signal.

To address these challenges, we propose \modelname{} (Fig.~\ref{fig:intro}(c)), a ULMM designed to think with visual updates during interleaved multimodal reasoning.
By
exploiting visual continuity across reasoning steps, \modelname{} incrementally predicts visual updates for each intermediate visual state rather than generating the full image.
Specifically, conditioned on historical visual representations, \modelname{}
models the modified content with a small set of visual update tokens, thereby avoiding redundant modeling of unchanged regions.
To
align token allocation with actual visual variation, \modelname{} introduces a temporal similarity (TSIM) router, which estimates the token budget required to preserve reconstruction quality under each level of visual change. To reduce
interference from
sample-level differences, we group samples with similar change magnitudes into the same
TSIM
interval. For each interval, we fit a
reconstruction-quality-token
curve and determine the token budget by stopping allocation once
the marginal gain in reconstruction quality falls below a predefined threshold. This strategy assigns fewer tokens to highly similar states and more tokens to complex visual transitions.

Training such a model further requires data that covers diverse reasoning processes and visual state updates.
However, existing interleaved multimodal reasoning datasets are often limited to specialized scenarios, such as mazes, puzzles, and visual search~\citep{li2025zebra,gu2025thinkmorph,li2025imagine,yang2026machine}, restricting their ability to support general multimodal reasoning.
To address this limitation, we construct StructCoT, a large-scale dataset with 1.05M samples spanning 44 task domains, enabling models to learn more generalizable reasoning
behaviors and visual-update patterns.

Our experiments show that, compared with the full-image generation paradigm, the visual-update paradigm provides a more effective way to support interleaved multimodal reasoning. When trained with StructCoT and large-scale multimodal data, \modelname{} further demonstrates strong performance across diverse multimodal reasoning and understanding tasks.

Our contributions are summarized as follows:

\begin{itemize}[leftmargin=*]

\item We propose \modelname{}, an update-centric
ULMM framework for interleaved multimodal reasoning. To the best of our knowledge, \modelname{} is the first
unified multimodal reasoning framework
that explicitly formulates interleaved multimodal reasoning as update-centric visual state modeling rather than repeated full-image generation. By representing sparse yet reasoning-critical visual changes with visual update tokens and adapting token allocation through the TSIM Router, \modelname{} reduces redundant visual modeling and concentrates supervision on visual state transitions.
\item We construct StructCoT, a large-scale interleaved multimodal reasoning dataset with 1.05M samples across 44 task domains.
StructCoT is organized around diverse reasoning structures and visual update patterns, providing broad supervision for generalizable update-centric multimodal reasoning.
\item
Extensive experiments
validate the effectiveness
of the visual-update paradigm.
\modelname{} reduces newly generated visual tokens by 55.6\% on average without compromising reconstruction fidelity and improves multimodal reasoning by 3.3\% over full-image generation.
Trained with StructCoT and large-scale multimodal data, \modelname{}
achieves a mean gain of 8.4\% over substantially larger open-source models on in-domain multimodal reasoning evaluations and improves over the comparable-scale Qwen3-VL-2B by 5.9\% on average across external multimodal reasoning and understanding benchmarks.
\end{itemize}

\section{Related Work}

\subsection{Interleaved Multimodal Reasoning}
\begin{figure*}[t!]
        \centering
        \includegraphics[width=\linewidth]{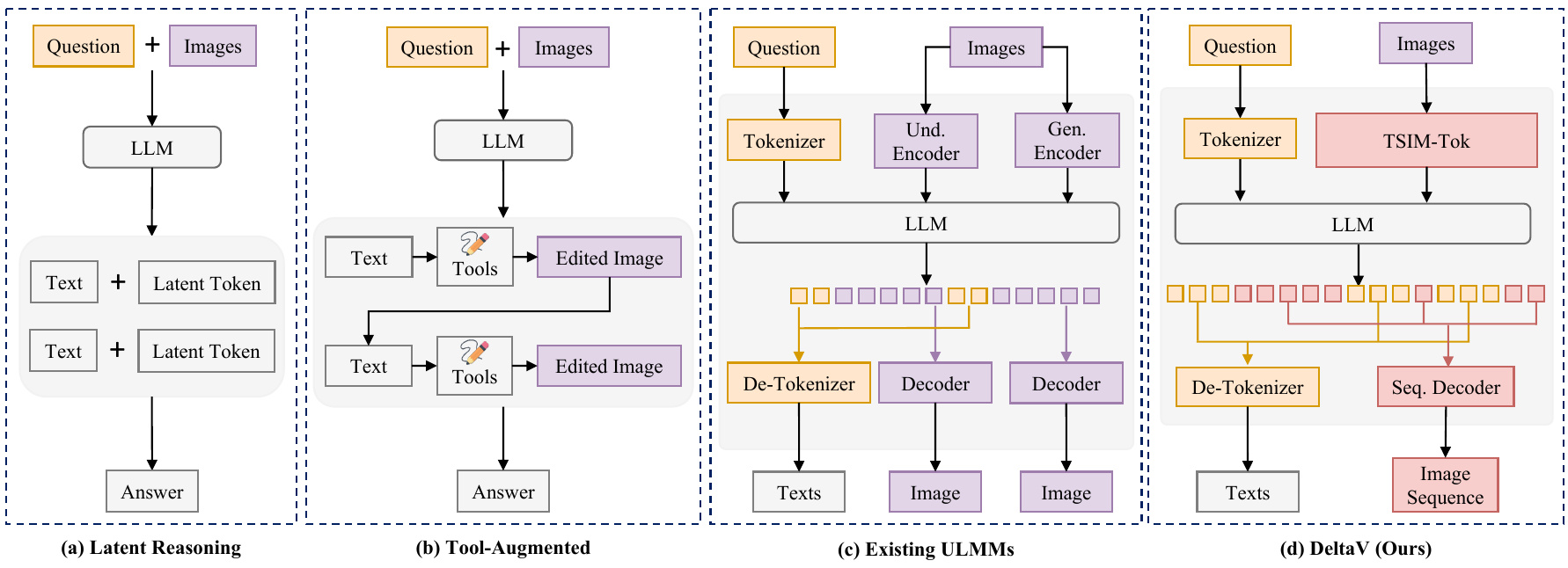}
        \caption{Existing interleaved multimodal reasoning methods include (a) latent reasoning, (b) tool-augmented reasoning, and (c) reasoning with existing ULMMs. Existing ULMMs rely on full-image generation for intermediate visual states, whereas (d) \modelname{} thinks with visual updates. By utilizing visual updates, \modelname{} increases visual information density and enhances multimodal reasoning. `Seq.’ denotes sequence.}
        \label{fig:related_work}
    \end{figure*}

Interleaved multimodal reasoning methods can be broadly categorized into implicit and explicit approaches. Implicit methods (Fig.~\ref{fig:related_work}(a)) typically compress visual information into perception tokens or latent visual representations, which reduces token overhead and strengthens text-based reasoning abilities~\citep{bigverdi2025perception, yu2025introducing, yang2026machine, wang2026monet, tong2026beyond}. While these methods are advantageous in controlling sequence length,
they sacrifice the fidelity and interpretability offered by explicit visual representations and remain fundamentally language-centric.
In contrast, explicit methods support reasoning by generating interpretable intermediate visual results and are often built upon tool-augmented, agent-based frameworks (Fig.~\ref{fig:related_work}(b))~\citep{gupta2023visual, zheng2025deepeyes, wang2025pixel, gao2025interleaved, menon2024whiteboard, zhou2025reinforced}. Representative works include visual programming with Python functions or dedicated vision modules~\citep{hu2024visual, gupta2023visual}, sketch generation for geometric, algorithmic, or spatial reasoning tasks~\citep{openai2025thinking, zhou2405image}, and the prediction of intermediate bounding boxes or structural annotations for fine-grained visual reasoning~\citep{chen2025sifthinker, zheng2025deepeyes}. Although these methods offer broad task coverage and rich expressiveness, their vision–language interactions typically rely on manually designed execution pipelines, making it difficult to achieve stable and generalizable interleaved reasoning.
Recent work on ULMMs that integrate multimodal understanding and generation has begun to explore interleaved text--image reasoning within a single model (Fig.~\ref{fig:related_work}(c))~\citep{zhao2025cot, shi2025mathcanvas, li2025zebra, gu2025thinkmorph, li2025imagine}.
However, under existing autoregressive or diffusion-based unified generation paradigms, modeling visual intermediate representations remains substantially more costly than
textual reasoning.
Consequently, this computational imbalance heavily restricts the scalability and efficiency of this framework for multi-step multimodal reasoning.

\subsection{Unified Large Multimodal Models}
Existing ULMMs can be broadly grouped into four paradigms.
The first paradigm appends diffusion-based image generation modules to the outputs of MLLMs~\citep{tong2025metamorph, lin2025uniworld, wu2025omnigen2}. Although this design enables limited synergy between visual understanding and generation~\citep{tong2025metamorph}, their interaction mainly occurs at the module level and lacks tight end-to-end coupling. As a result, visual generation cannot be continuously involved in the multimodal reasoning process.
The second paradigm adopts autoregressive modeling to jointly generate text and images~\citep{chen2025janus, fang2025puma, han2412infinity, wu2026liquid}. Its main limitation is that the number of image tokens grows
rapidly with image resolution.  For example, the model in~\citet{chen2025janus} requires 576 tokens to generate a single image at a resolution of
384x384, while~\citet{han2412infinity} exceeds 10,000 tokens at a
1024x1024. In contrast, a single step of text reasoning typically requires only 50 to 100 tokens. This severe imbalance in sequence length causes reasoning signals to be dominated by image tokens, which degrades overall reasoning performance.
The third paradigm combines autoregressive text generation with diffusion-based image generation in a hybrid framework~\citep{deng2025emerging, he2025emma, liao2025mogao}.
Although text is produced by standard single-pass autoregressive decoding, high-quality image generation requires dozens of diffusion denoising steps.
This results in strong asymmetry between text and image generation in both computational cost and reasoning pace, making it inefficient for multi-step reasoning.
The fourth paradigm consists of recently proposed fully diffusion-based unified generation and understanding models~\citep{yang2025mmada, li2025lavida, xin2025lumina}. While these methods achieve strong image generation quality, their semantic understanding and explicit reasoning abilities remain limited,
making them less suitable
for reasoning-intensive multimodal tasks.

Overall, existing interleaved multimodal reasoning methods face a trade-off between efficiency and reasoning fidelity,
while ULMMs remain constrained by inefficient visual information modeling.
By generating each intermediate visual state as a full image, these methods repeatedly model largely unchanged content, introducing redundant visual supervision, diluting the learning signal for reasoning-critical state transitions, and
weakening cross-step consistency.
In contrast, \modelname{} is a ULMM designed to enhance visual information density by exploiting structural continuity through variation-aware visual updates and dynamic token allocation, supporting efficient and effective interleaved multimodal reasoning.

\section{\modelname{}}

\subsection{Unified Autoregressive Modeling with Visual Updates}

Given a reasoning trajectory, conventional autoregressive ULMMs serialize interleaved multimodal reasoning as a sequence of text tokens and complete visual state tokens:
\begin{equation}
Y = \{Z_0, X_1, Z_1, X_2, Z_2, \dots \},
\end{equation}
where $X_t$ denotes the text tokens at step $t$, and $Z_t$ denotes the complete visual representation of the intermediate image $I_t$.
This formulation requires the model to repeatedly predict full visual states, although consecutive intermediate images often share substantial visual content.

\modelname{} reformulates this process by decomposing the visual trajectory into an initial visual state and a sequence of visual updates:
\begin{equation}\label{eq:2}
Y = \{Z_0, X_1, \Delta Z_1, X_2, \Delta Z_2, \dots \},
\end{equation}
where $Z_0$ represents the initial visual state tokens, and $\Delta Z_t$ denotes variable-length visual update tokens for step $t$.
The length of $\Delta Z_t$ is adaptively determined by the degree of visual change.
This formulation enables the language model to reason autoregressively over text and compact visual updates, while avoiding repeated prediction of complete visual states.
This decomposition shifts the learning target from state reconstruction to transition modeling: instead of repeatedly predicting visual content that is already explained by previous states, the model is supervised to capture the step-specific visual evidence responsible for moving the reasoning trajectory forward.

\begin{figure*}[t!]
        \centering
        \includegraphics[width=1\linewidth]{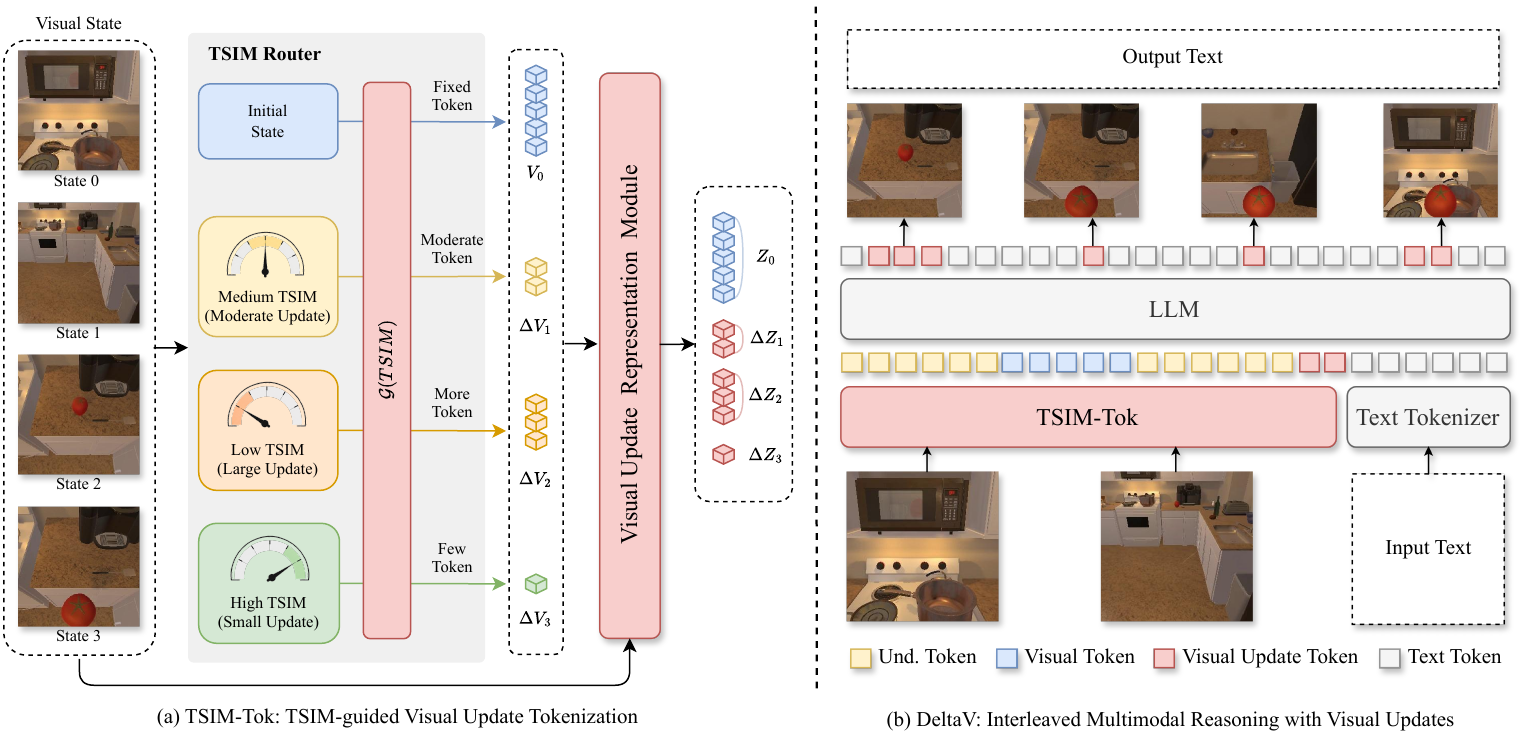}
        \caption{Overview of \modelname{}. \modelname{} represents visual reasoning as a base visual state followed by incremental visual updates. TSIM-Tok produces variation-aware update tokens, and the TSIM Router assigns adaptive token budgets based on temporal visual variation, where lower TSIM indicates larger changes. This allows the LLM to alternate between text and visual updates while focusing visual modeling on reasoning-relevant changes. `Und.' denotes understanding.}
        \label{fig:method}
    \end{figure*}

As shown in Fig.~\ref{fig:method}, \modelname{} implements this formulation with TSIM-Tok, which converts visual observations into a base token set and variation-aware visual update tokens.
Through this update-centric tokenization, \modelname{} allocates fewer tokens to minor updates and more tokens to complex visual transitions, thereby reducing redundant visual modeling while preserving reasoning-relevant visual changes.
The detailed token allocation mechanism and its use in interleaved reasoning are introduced in the following sections.

\subsection{TSIM-Tok: Variation-Aware Tokenization for Visual Updates}
\label{sec:TSIM-Tok}
TSIM-Tok serves as the visual tokenizer that supports variation-aware visual updates in \modelname{}. It is designed to address three key questions: how to estimate the degree of visual change along a reasoning trajectory, how to determine an appropriate token budget according to such variation, and how to encode each visual state as compact base or visual update tokens under the allocated budget. We first introduce the TSIM Router for dynamic token allocation, and then introduce the Visual Update Representation Module for encoding visual states under the allocated budgets.

\paragraph{TSIM Router.}

In general, fewer new visual details require fewer visual update tokens, while complex visual updates demand more. Based on this idea, we propose the
temporal similarity (TSIM) router, which is a
variation-driven dynamic token allocation mechanism that maps temporal visual change to an adaptive update-token budget.

To quantify the variation of each visual state relative to its historical context, we define temporal similarity.
Specifically, for the visual image at step $t$, we first extract patch-level visual features using a visual feature extractor $\mathcal{E}$:
\begin{equation}
\mathbf{F}'_t = \mathcal{E}(I_t).
\end{equation}
Then, we compute cosine similarities between the current feature $\mathbf{F}'_t$ and all preceding features in the sequence. The pairwise similarity between steps $t$ and $j$ is defined as:
\begin{equation}
s_{tj} =
\frac{
\mathbf{F}'_t \cdot \mathbf{F}'_j
}{
\|\mathbf{F}'_t\|_2 \, \|\mathbf{F}'_j\|_2
}, \quad j < t.
\end{equation}
Intuitively, historical visual states should not contribute equally to the temporal similarity of the current state.
We weight each preceding state by two factors: temporal proximity and information capacity.
Specifically, we introduce a temporal decay coefficient $\alpha \in [0,1]$ to emphasize recent states, which are typically more correlated with the current observation, while reducing the influence of outdated historical information.
Meanwhile, the token budget $K_j$ serves as an indicator of the information capacity of state $j$, as states assigned more tokens are expected to preserve richer visual information and thus provide more informative references.
Accordingly, we define TSIM as:
\begin{equation}
\label{tsim}
\mathrm{TSIM}_t =
\frac{
\sum_{j=0}^{t-1} s_{tj} \cdot \alpha^{t-1-j} \cdot K_j
}{
\sum_{j=0}^{t-1} \alpha^{t-1-j} \cdot K_j
}.
\end{equation}

    \begin{figure*}[t!]
        \centering
        \includegraphics[width=1\linewidth]{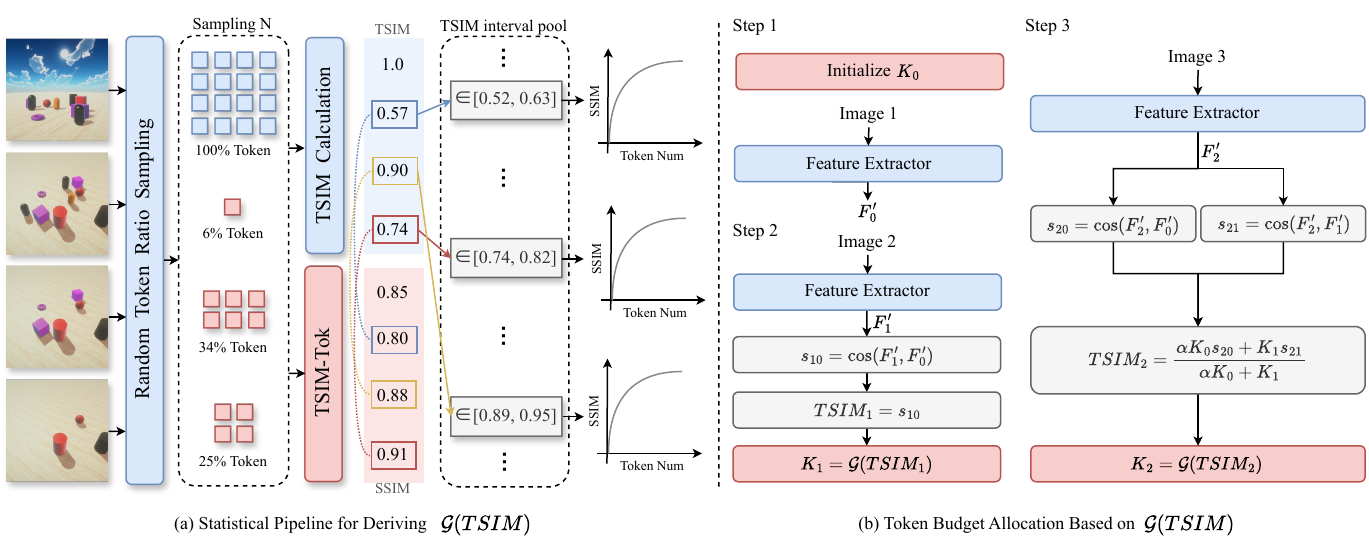}
        \caption{TSIM-driven Token Allocation Pipeline.
        (a) Offline calibration for estimating the relationship among TSIM, token budget, and reconstruction fidelity.
        (b) Token budget allocation based on the calibrated TSIM-to-budget mapping.}
        \label{fig:sim_sample}
    \end{figure*}

To convert TSIM into an actionable allocation rule, we need to answer a key question:
given a certain degree of temporal visual variation, how many visual update tokens are sufficient to preserve the current visual state?
To this end, we perform an offline calibration procedure that empirically estimates the relationship among TSIM, token budget, and reconstruction fidelity.

The pipeline is illustrated in Fig.~\ref{fig:sim_sample}(a).
Given an image sequence $\{I_0, I_1, \ldots, I_T\}$, we randomly sample a visual update token budget $K_t$ for each visual state.
For each sampled $K_t$, we compute the corresponding $\mathrm{TSIM}_t$ using Eq.~\ref{tsim}.
In parallel, we reconstruct $I_t$ with a pretrained variable-token TSIM-Tok under the budget $K_t$, and measure the reconstruction quality using SSIM.
We repeat this process $N$ times to expose each visual state to different compression levels and observe how reconstruction quality varies with the token budget, yielding a set of calibration samples:
\begin{equation}
\{(\mathrm{TSIM}_t^{(n)}, K_t^{(n)}, \mathrm{SSIM}_t^{(n)})\}_{n=1}^N.
\end{equation}

Then, we partition the TSIM range into $M$ disjoint intervals $\{\mathcal{B}_m\}_{m=1}^{M}$ to group samples with similar temporal variation.
Within each interval, the effect of TSIM is locally controlled, so the remaining variation in SSIM can be more reliably attributed to the token budget, leading to a more stable estimate of the token--fidelity relationship.
Accordingly, we aggregate samples by token budget within each interval and compute the empirical expected reconstruction quality:
\begin{equation}
E_m(K) = \mathbb{E}\left[ \mathrm{SSIM} \mid K_t = K, \mathrm{TSIM}_t \in \mathcal{B}_m \right],
\quad K \in \mathcal{K},
\end{equation}
where $\mathcal{K}$ denotes the set of candidate token budgets considered in the offline sampling process.

Since $E_m(K)$ is observed only on the discrete set $\mathcal{K}$, it does not directly characterize how reconstruction quality changes between adjacent or unseen token budgets.
We fit a continuously differentiable curve to the empirical observations in each TSIM interval:
\begin{equation}
C_m(x) = \mathrm{CurveFit}\big(\{ (K, E_m(K)) \mid K \in \mathcal{K} \}\big),
\end{equation}
where $x \in \mathbb{R}^+$ denotes a continuous token budget variable.
The fitted curve $C_m(x)$ provides a smooth approximation of how reconstruction fidelity changes with token allocation under the corresponding TSIM interval $\mathcal{B}_m$.

We use the fitted token--fidelity curve $C_m(x)$ to determine the token budget at which additional tokens begin to yield diminishing returns.
Specifically, the derivative $\frac{\mathrm{d} C_m(x)}{\mathrm{d} x}$ measures the marginal improvement in reconstruction quality obtained by increasing the token budget.
A large derivative indicates that additional tokens still provide noticeable reconstruction gains, whereas a small derivative suggests that the reconstruction quality has started to saturate.
Given a predefined threshold $\tau > 0$, we identify the budget at which the marginal gain falls below $\tau$, and then choose the closest value in the discrete candidate set:
\begin{equation}
K_t = \mathcal{G}(\mathrm{TSIM}_t) =
\arg\min_{x \in \mathcal{K}}
\left| x - x_m^\star \right|,
\quad
x_m^\star = \inf\left\{
z \,\bigg|\,
\frac{\mathrm{d} C_m(z)}{\mathrm{d} z} \le \tau
\right\},
\quad
\mathrm{TSIM}_t \in \mathcal{B}_m .
\end{equation}

This slope-based criterion allocates tokens only while they are expected to provide sufficient improvement for reconstructing the current visual state.
Once the marginal gain falls below $\tau$, further increasing the token budget is considered inefficient, and the corresponding discrete budget is selected.
Accordingly, the offline calibration procedure establishes the TSIM Router allocation rule $\mathcal{G}$, which maps each temporal similarity score to an appropriate visual update token budget according to its TSIM interval.

In practice, the TSIM Router is used to pre-allocate token budgets for images within interleaved multimodal reasoning chains during training, as illustrated by the example of the first three images in Fig.~\ref{fig:sim_sample}(b).
This supervision enables \modelname{} to learn the relationship between visual variation and token demand. During inference, \modelname{} autoregressively generates visual update tokens and automatically stops when sufficient visual information has been produced.

\paragraph{Visual Update Representation Module.}
After obtaining the dynamic token budget from the TSIM Router, we use the Visual Update Representation Module to instantiate visual queries with step-specific lengths and convert each visual state into base or visual update tokens. This module consists of a visual backbone, learnable base and update queries, update aggregation layers, quantization-related modules, and lightweight generation heads. The visual backbone extracts image features for each visual state, while the newly introduced update components encode these features into base tokens and variable-length visual update tokens.
As shown in Fig.~\ref{fig:method}, given a reasoning image sequence
$\{I_0, I_1, \ldots, I_T\}$, the TSIM Router assigns a visual update token budget $K_t=\mathcal{G}(\mathrm{TSIM}_t)$ for each state $I_t$ with $t>0$.
Based on these budgets, we parameterize the visual queries as
\begin{equation}
V = \{V_0, \Delta V_1, \Delta V_2, \ldots, \Delta V_T\},
\end{equation}
where $V_0 \in \mathbb{R}^{K_0 \times d}$ denotes the base queries for the initial visual state $I_0$, and
$\Delta V_t \in \mathbb{R}^{K_t \times d}$ denotes the visual update queries for step $t$.
Here, $K_0$ is the base token budget, $K_t$ is the dynamic visual update budget allocated by the TSIM Router, and $d$ is the query dimension.

The base queries $V_0$ interact with the initial image $I_0$ to produce the base visual token set $Z_0$, which serves as the initial visual representation.
For each subsequent state $I_t$, the visual update queries $\Delta V_t$ are conditioned on the initial visual tokens and preceding visual updates, producing the corresponding visual update token set $\Delta Z_t$:
\begin{equation}
Z_0 = \Phi(I_0, V_0),
\qquad
\Delta Z_t = \Phi(I_t, \Delta V_t \mid Z_0, \Delta Z_{<t}), \quad t=1,\ldots,T,
\end{equation}
where
$\Delta Z_{<t}=\{\Delta Z_1,\ldots,\Delta Z_{t-1}\}$
denotes the historical visual updates before step $t$, and $\Phi(\cdot)$ denotes the visual update representation module.
This sequential notation highlights the temporal dependency among visual updates, while the corresponding dependency structure is implemented with causal attention masks inside $\Phi(\cdot)$.
Therefore, all visual states and parameterized queries can be processed jointly without allowing the tokenization of $I_t$ to access future states.
In this way, TSIM-Tok preserves stable visual content through historical tokens, while allocating new tokens only to visual updates that cannot be sufficiently explained by previous states.
The detailed architecture
of TSIM-Tok
is provided in App.~\ref{app:visual_update_module}.

\subsection{Interleaved Multimodal Reasoning with Visual Updates}
\label{sec:interleaved_reasoning_with_updates}

We now describe how the visual update formulation is used for interleaved multimodal reasoning.
At the LLM level, \modelname{} follows the sequence in
Eq.~\ref{eq:2}, where the initial visual state is represented by $Z_0$, and each subsequent reasoning step generates only the newly required visual update tokens $\Delta Z_t$ after the textual reasoning tokens $X_t$.
Unlike
full-image modeling, \modelname{} does not ask the LLM to regenerate a complete visual state $Z_t$ at every step.
Although historical visual tokens are used inside TSIM-Tok to condition each visual update,
only the newly generated update tokens are appended to the LLM sequence at each step.
As a result, the autoregressive sequence is built around newly generated visual updates, which keeps visual reasoning compact while preserving the temporal structure of the reasoning process.

During training, the TSIM Router assigns a token budget to each visual update according to temporal visual variation, and TSIM-Tok encodes the corresponding visual state into a variable-length update sequence.
To make this budget decision learnable by the autoregressive model, we append an explicit \texttt{\textless{}|vision\_end|\textgreater{}} token after each visual update sequence as boundary supervision.
In this way, the model learns both what visual information to generate and when the current update should stop.
Therefore, TSIM serves as a teacher policy during training, providing supervision for learning variation-aware visual update generation.

Since the target future visual state is unavailable during inference, \modelname{} cannot directly compute its TSIM score and therefore does not perform online TSIM estimation or explicit budget prediction.
Instead, after each textual reasoning step, the model autoregressively generates visual update tokens until it emits \texttt{\textless{}|vision\_end|\textgreater{}}.
This learned stopping mechanism transfers the TSIM-supervised update-length policy to inference, allowing \modelname{} to adaptively determine the length of each visual update without repeatedly generating full images.

\section{Dataset}

\modelname{} models visual reasoning as a progressive update process, where visual states are continuously refined throughout the reasoning procedure. Training such a model requires supervision over diverse reasoning processes and intermediate visual representations. However, existing interleaved multimodal reasoning datasets provide only limited coverage.  Zebra-CoT~\citep{li2025zebra} contains approximately 180K samples across 21 task domains, while other related datasets~\citep{gu2025thinkmorph,li2025imagine,yang2026machine} mainly focus on specialized scenarios such as mazes, puzzles, or visual search.
Although these datasets are valuable for specific reasoning settings,
their limited scale and diversity constrain the variety of supervision available for training general multimodal reasoning models.

\paragraph{StructCoT Dataset.}
    \begin{figure*}[t!]
        \centering
        \includegraphics[width=1\linewidth]{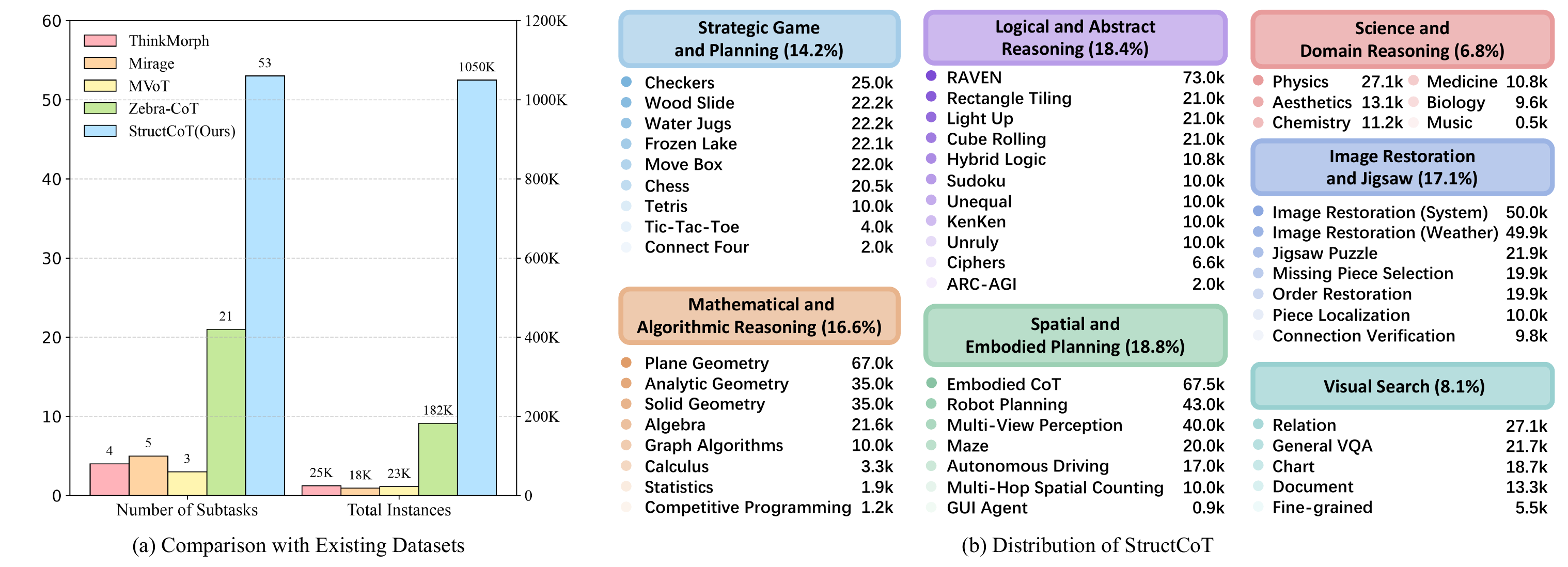}
        \caption{Comparison of StructCoT with existing interleaved multimodal reasoning datasets and detailed task classification.}
        \label{fig:StructCoT}
    \end{figure*}

To overcome these limitations, we construct StructCoT, a large-scale interleaved multimodal reasoning dataset with 1.05M samples spanning 44 task domains. Compared with Zebra-CoT, StructCoT is approximately 5.8$\times$ larger in scale and more than doubles the number of task domains. The dataset contains a broad range of intermediate visual representations, including grids, graphs, trajectories, viewpoint transformations, and state diagrams, which are progressively updated throughout the reasoning process. This coverage introduces diverse visual state transitions and reasoning trajectories that are underrepresented in existing datasets. Consequently, StructCoT offers more comprehensive supervision for learning generalizable multimodal reasoning behaviors.

\begin{figure*}[t!]
        \centering
        \includegraphics[width=0.9\linewidth]{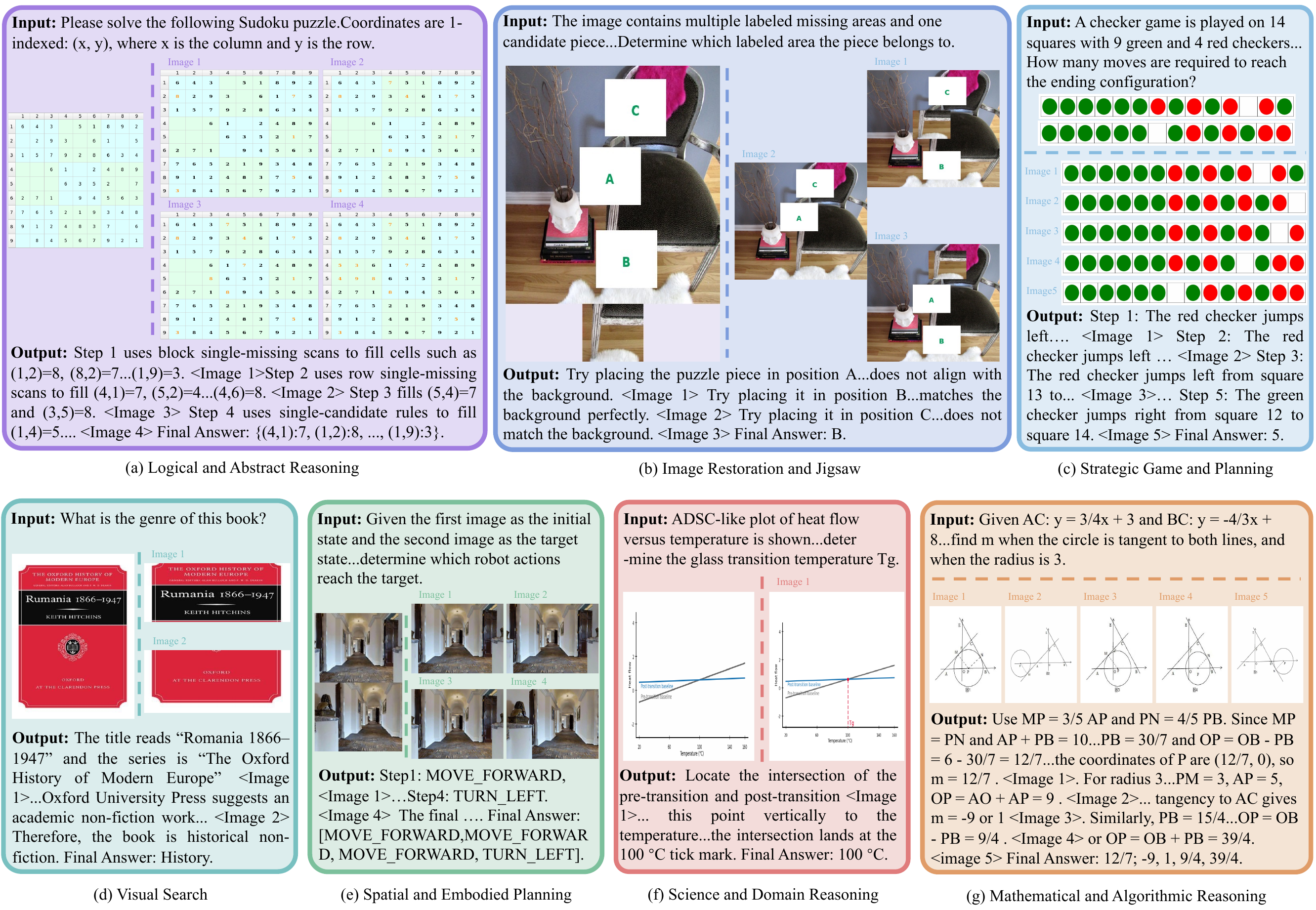}
        \caption{Examples of interleaved textual and visual reasoning processes across the seven task categories in StructCoT.}
        \label{fig:struct_example}
    \end{figure*}

\paragraph{Task Categorization by Reasoning Structure.}
To systematically cover different forms of visual state evolution, StructCoT categorizes tasks according to the transition patterns of intermediate visual states and the computational characteristics of their reasoning processes.
As illustrated in Fig.~\ref{fig:StructCoT}, StructCoT organizes tasks into seven representative reasoning categories: strategic planning, spatial reasoning, logical reasoning, mathematical reasoning, scientific reasoning, visual search, and image reconstruction.
Representative examples are shown in Fig.~\ref{fig:struct_example}, with detailed descriptions provided in App.~\ref{appendix:dataset}.

Overall, this reasoning-structure-driven taxonomy is designed to cover diverse visual evolution patterns, thereby supporting the learning of generalizable visual updates across interleaved multimodal reasoning tasks.

\section{Experiments}
\subsection{Implementation}
\paragraph{Training Configuration.}
\modelname{} adopts SigLIP2-Large~\citep{tschannen2025siglip2} as the frozen visual backbone for TSIM-Tok and a Qwen3-based~\citep{Qwen3-VL} LLM for language modeling.
The remaining visual-update components in TSIM-Tok are newly introduced and trained within our framework.
Our training procedure consists of two stages. We first train the visual tokenizer TSIM-Tok, and then freeze it while training \modelname{} on top of the learned visual representations.

Specifically, TSIM-Tok is trained progressively with single-image and multi-image data. Its training corpus includes 60M image understanding samples, 40M image generation samples, and the training split of StructCoT.
During the entire training process, the visual backbone remains frozen, while the newly introduced components learn temporally conditioned visual updates.

After TSIM-Tok training is completed, we freeze the tokenizer and train \modelname{} through an alignment stage followed by supervised fine-tuning. The training data includes 25M image understanding samples and 10M interleaved multimodal reasoning samples, supporting cross-modal alignment and interleaved reasoning. Detailed training configurations and data compositions are provided in App.~\ref{Training Details} and Tab.~\ref{tab:training_config}.

\paragraph{Evaluation Setup.} Zebra-CoT\footnote{We use the Zebra-CoT dataset from \url{https://huggingface.co/datasets/multimodal-reasoning-lab/Zebra-CoT}, which is released under the CC BY-NC 4.0 license.
Our use of this dataset is limited to academic research purposes and does not involve any commercial activities.} is a large-scale multimodal reasoning dataset spanning 21 task domains and comprising 180K samples. The dataset is partitioned into training, validation, and test splits using task-stratified sampling with an 8:1:1 ratio.
For image reconstruction ablation experiments, models are evaluated on the test split.
For multimodal reasoning ablation experiments, we randomly sample 4K examples from the test split for evaluation.

For a more comprehensive evaluation of multimodal reasoning performance, we construct a StructCoT test set by randomly sampling 800 examples from each of the seven major task categories, yielding a total of 5.6K test samples. To comprehensively assess \modelname{}'s generalization ability, we evaluate it on a diverse suite of external benchmarks covering multimodal reasoning and understanding. Details of
the evaluation pipeline and benchmarks
are provided in
App.~\ref{appendix:Evaluation_Details} and App.~\ref{appendix:benchmark_details}.

    \begin{figure*}[t!]
        \centering
        \includegraphics[width=1\linewidth]{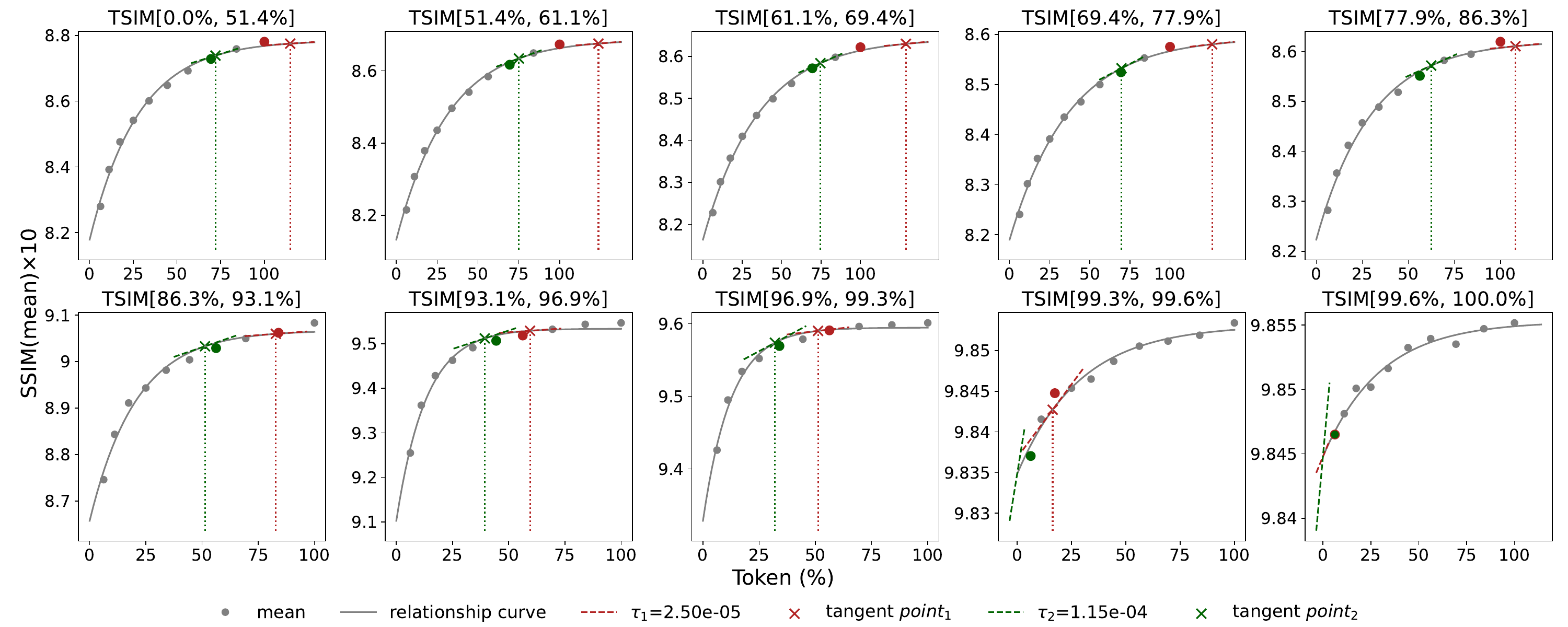}
        \caption{Derivation of the TSIM Router allocation rule. For each TSIM interval, we fit the relationship between the visual update token budget and reconstruction fidelity, and select the budget near the point where the marginal reconstruction gain falls below the threshold $\tau$.}
        \label{fig:fit_tangent}
    \end{figure*}

\subsection{Deriving the TSIM Router Allocation Rule}
\label{sec:token-count-similarity}

We derive the concrete allocation rule $\mathcal{G}$ through an offline calibration procedure.
We choose offline calibration rather than directly learning token allocation end-to-end because it provides an interpretable and stable budget-control mechanism, while avoiding optimization complexity from jointly learning representation, allocation, and stopping decisions.
Given the TSIM definition in Sec.~\ref{sec:TSIM-Tok}, the goal is to determine an appropriate visual update token budget for each TSIM level based on marginal reconstruction gains. This analysis is performed on the Zebra-CoT validation split, which contains 18K multimodal reasoning image sequences.

To support this analysis, we first train a TSIM-Tok model on the Zebra-CoT training set. The model supports visual updates with 10 predefined token budgets, forming the candidate set $\mathcal{K}=\{9,16,25,36,49,64,81,100,121,144\}$. The initial visual state uses a fixed budget of $K_0=144$, while the budgets of subsequent visual updates are selected from $\mathcal{K}$.
For a sequence containing $L+1$ images (i.e., $L$ visual update steps), we sample $N = n\cdot L^2$ configurations, where $n=|\mathcal{K}|=10$.
This assigns more evaluations to longer sequences, improving curve estimation while keeping the computational cost manageable.
For each sampled configuration, we compute its TSIM score according to Eq.~\ref{tsim}.
Here, $\mathcal{E}$ is instantiated as DINOv2~\citep{oquab2023dinov2} for offline temporal similarity estimation, since its self-supervised features provide robust semantic visual representations for image similarity and semantic correspondence~\citep{zhang2023tale,huber2025synthetic}.
TSIM-Tok itself
instead adopts SigLIP2-Large~\citep{tschannen2025siglip2} as its visual backbone, which provides language-aligned visual representations for unified multimodal understanding and generation, consistent with recent semantic-aware unified visual tokenizers~\citep{zheng2026vision,du2026vqrae}.
Specifically, we feed the image sequence into TSIM-Tok with the sampled token budgets, reconstruct the visual states using its lightweight reconstruction decoder, and measure the reconstruction quality using SSIM.

In the analysis stage, we divide samples into $M$ TSIM intervals with approximately equal sample counts. We choose $M = 10$ as the default setting because it provides a stable trade-off between modeling granularity and statistical reliability. We also find that the allocation rule is robust to the number of intervals, with detailed ablations provided in App.~\ref{appendix:interval_ablation}. Within each interval, we fit the relationship between token count and SSIM.
The overall results are illustrated in Fig.~\ref{fig:fit_tangent}.
As the number of tokens increases, the SSIM improvement gradually saturates. We use a slope threshold $\tau$ to control the visual-update token budget, and select the preset budget closest to the corresponding tangent point in each TSIM interval. Here, $\tau$ serves as a budget-control parameter, enabling comparisons between full-image modeling and visual updates under controlled reconstruction-quality and token-budget settings.
The resulting TSIM-to-budget mapping is then used in the subsequent training of \modelname{}: each training sample is pre-assigned a visual update token budget according to its TSIM value, enabling the model to learn variation-aware token allocation across images with different degrees of visual change.

\subsection{Analysis of Visual Updates}

    \begin{figure*}[t!]
        \centering
        \includegraphics[width=1\linewidth]{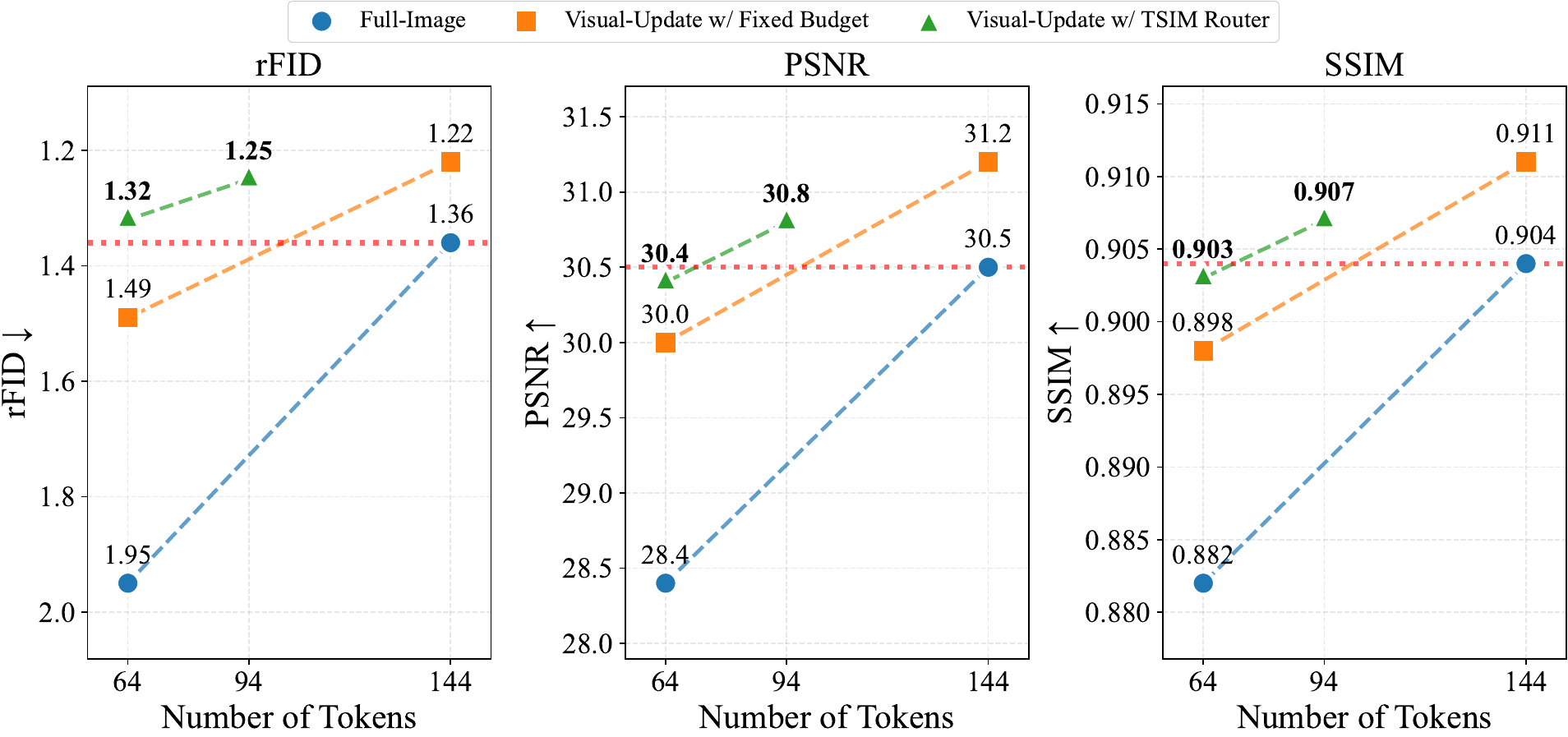}
        \caption{Comparison of full-image modeling and visual-update modeling for image reconstruction under different token budgets. ``Full-Image'' denotes full-image modeling, while ``Visual-Update'' denotes visual-update modeling. We set the maximum number of visual update tokens to 144, so the average number of tokens used by Visual-Update w/ TSIM Router always remains below 144.}
        \label{fig:encoding_comparison}
    \end{figure*}

\subsubsection{Do Visual Updates Improve Reconstruction Efficiency?}

In this section, we isolate the effect of visual updates by comparing settings that use the same model architecture and training data, but differ in the visual modeling strategy.
We first analyze how \modelname{}'s visual updates improve image reconstruction efficiency compared with full-image modeling. This reconstruction study focuses on TSIM-Tok, the visual tokenizer used by \modelname{} to encode visual updates.
We consider two settings: (1) visual updates with fixed token budget, and (2) TSIM-Router-driven visual updates, which dynamically allocate tokens following Sec.~\ref{sec:token-count-similarity}. As shown in Fig.~\ref{fig:encoding_comparison}, visual updates with fixed token budget consistently achieve better reconstruction quality than full-image modeling under the same average token count.
We hypothesize that this improvement is partly due to the reuse of historical visual representations,
which reduces redundant modeling of unchanged visual content. As a result, under a constrained token budget, visual updates may focus more on modified visual information at the current step.

TSIM-Router-driven visual updates further improve reconstruction quality, especially in token-limited settings (64 tokens).
Beyond improvements under the same token budgets, TSIM-Router-driven visual updates also demonstrate higher token efficiency.
Specifically, using only 64 visual update tokens per step on average, TSIM-Router-driven visual updates achieve reconstruction quality comparable to full-image modeling with 144 visual tokens.
This indicates that, while preserving comparable TSIM-Tok reconstruction fidelity, \modelname{} reduces the number of newly generated visual tokens appended to the LLM sequence by 55.6\% on average.
We believe this gain comes from the TSIM-Router-driven strategy, which enables the model to actively
Discard redundant content and learn a more change-sensitive allocation behavior. When the average token count increases to 94, TSIM-Router-driven visual updates surpass the full-image modeling configuration with 144 tokens across all reconstruction metrics. This result suggests that additional tokens are better allocated to fine-grained visual variations, rather than uniformly modeling redundant content as in full-image modeling. Such a property provides a foundation for shifting optimization away from redundant low-level reconstruction and toward reasoning-relevant visual state updates in subsequent multi-step reasoning tasks.

\begin{table*}[t]
\centering
\caption{\textbf{Effectiveness of \modelname{}'s visual updates in multimodal reasoning.}
}\label{tab:acc_ablation}

\begin{adjustbox}{max width=0.98\linewidth}
\begin{threeparttable}
\begin{tabular}{l|ccccc}
\specialrule{.2em}{.1em}{.1em}

\multicolumn{1}{c|}{Setting}  & \hspace{0.2cm}2D Reasoning\hspace{0.2cm}        & \hspace{0.2cm}3D Reasoning\hspace{0.2cm}      & \hspace{0.2cm}Scientific Reasoning\hspace{0.2cm}  & \hspace{0.2cm}Strategic Reasoning\hspace{0.2cm}  & \hspace{0.2cm}Overall\hspace{0.2cm} \\
\hline
Text-Only         & 50.4   & 55.1    & 43.6  &  40.1 & 47.3      \\
Full-Image 144        & 46.4   & 50.9    & 43.7  &  40.3 & 45.3      \\
Visual-Update w/ Fixed Budget 144  & \textbf{50.8}    & 51.5          & 43.5    & 42.0    & 47.0 \\
Visual-Update w/ TSIM Router 64  & 48.8    & \textbf{56.0}          & \textbf{45.9}    & \textbf{43.7}   & \textbf{48.6} \\
\specialrule{.2em}{.1em}{.1em}

\end{tabular}
\begin{tablenotes}[para,flushleft]
\footnotesize
All models are trained and evaluated on the Zebra-CoT dataset. ``Full-Image'' denotes full-image modeling, while ``Visual-Update'' denotes visual-update modeling.

\end{tablenotes}
\end{threeparttable}
\end{adjustbox}
\end{table*}

\subsubsection{Do Visual Updates Improve Interleaved Multimodal Reasoning?}
\label{para:inc_vcot}
After establishing that \modelname{}’s visual updates achieve comparable reconstruction quality to full-image modeling with fewer tokens, we further evaluate their impact on multimodal reasoning.
Specifically, we conduct experiments on the Zebra-CoT dataset under four settings: (1) text-only reasoning; (2) interleaved reasoning with 144 tokens using full-image modeling; (3) interleaved reasoning with fixed 144 tokens using visual updates; and (4) interleaved reasoning with an average of 64 tokens using TSIM-Router-driven visual updates.

As shown in Tab.~\ref{tab:acc_ablation}, full-image modeling performs worse than text-only reasoning, indicating that naively introducing intermediate image generation does not necessarily improve multimodal reasoning.
This supports our motivation that the bottleneck of full-image generation lies not only in its token cost, but also in its supervision structure.
Since consecutive visual states are often highly correlated, full-image modeling repeatedly supervises the reconstruction of largely unchanged and task-irrelevant content, which can bias optimization toward low-level reconstruction and weaken the learning signal on sparse but reasoning-critical visual transitions.

In contrast, fixed-budget visual updates outperform full-image modeling across most subtasks, improving the overall score by 1.7\%.
This suggests that modeling visual updates provides a denser reasoning-relevant learning signal under the same token budget.
However, its performance remains slightly below text-only reasoning, suggesting that a fixed budget still leaves redundancy in visually minor transitions.

Finally, TSIM-Router-driven visual updates achieve the best overall performance, surpassing text-only reasoning by 1.3\% and improving over full-image modeling by 3.3\%, while using only 64 visual update tokens on average.
Notably, this setting attains reconstruction quality comparable to full-image modeling with 144 tokens, suggesting that the gain does not simply come from better reconstruction fidelity.
Instead, by adapting the token budget to the magnitude of visual change, the TSIM Router further reduces redundant visual modeling and increases the supervision density of reasoning-relevant visual transitions.
These results suggest that \modelname{} mitigates the negative effect of naive intermediate image generation and turns visual updates into a more compact and useful reasoning signal. Additional analyses with different generated visual-token budgets are provided in App.~\ref{appendix:tau_sensitivity}, where the 64-token setting shows the best overall trade-off.

However, routed visual updates do not surpass the text-only baseline on 2D reasoning tasks (48.8 vs. 50.4). Increasing the budget to a fixed 144 tokens marginally surpasses text-only performance (50.8 vs. 50.4), confirming that preserving local details requires higher token capacity.

\subsubsection{Do Visual Updates Improve General Multimodal Understanding and Reasoning?}

We further examine whether visual-update training improves general multimodal understanding and reasoning beyond the in-domain interleaved reasoning setting. This experiment is designed as a controlled ablation rather than a comprehensive benchmark comparison: both models are initialized from the same baseline, trained on the same StructCoT data for the same number of steps, and differ only in whether intermediate visual states are retained. For the text-only baseline, we remove intermediate images while preserving the original reasoning texts and logical structures.

As shown in Tab.~\ref{tab:multi_benchmark}, visual-update training
improves over the text-only baseline on most evaluated understanding and reasoning benchmarks. These results indicate that visual updates provide complementary supervision beyond textual reasoning traces, helping the model learn more transferable multimodal representations.

\begin{table*}[t]
\centering
\caption{\textbf{Controlled ablation between text-only training and \modelname{}'s visual-update training.}}
\label{tab:multi_benchmark}

\begin{adjustbox}{max width=0.98\linewidth}
\begin{threeparttable}
\begin{tabular}{l|ccccccc}
\specialrule{.2em}{.1em}{.1em}

\multicolumn{1}{c|}{Setting}
& \hspace{0.15cm}MMVP\hspace{0.15cm}
& \hspace{0.15cm}MM-Vet\hspace{0.15cm}
& \hspace{0.15cm}MathVista\hspace{0.15cm}
& \hspace{0.15cm}VisuLogic\hspace{0.15cm}
& \hspace{0.15cm}LogicVista\hspace{0.15cm}
& \hspace{0.15cm}BLINK\hspace{0.15cm}
& \hspace{0.15cm}EMMA\hspace{0.15cm}  \\

\hline

Text-Only
& 48.0   & 47.3   & 66.0
& 24.8   & 35.8   & 48.9
& 23.9   \\

Visual-Update w/ TSIM Router
& 48.0
& \textbf{51.6\textcolor{darkgreen}{(4.3$\uparrow$)}}
& \textbf{67.1\textcolor{darkgreen}{(1.1$\uparrow$)}}
& \textbf{25.5\textcolor{darkgreen}{(0.7$\uparrow$)}}
& \textbf{36.3\textcolor{darkgreen}{(0.5$\uparrow$)}}
& \textbf{52.0\textcolor{darkgreen}{(3.1$\uparrow$)}}
& \textbf{25.9\textcolor{darkgreen}{(2.0$\uparrow$)}}  \\

\specialrule{.2em}{.1em}{.1em}

\end{tabular}

\begin{tablenotes}[para,flushleft]
\footnotesize
All models are initialized from the same baseline and trained for 30,000 steps on the same dataset.
\end{tablenotes}

\end{threeparttable}
\end{adjustbox}
\end{table*}

\subsubsection{Ablation on Internal Components of \modelname{}}

\paragraph{TSIM-driven token allocation ablation.}

To validate the effectiveness of TSIM-driven dynamic token allocation, we compare different modeling strategies under a comparable average token budget of about 64 tokens. As shown in Tab.~\ref{tab:recon_ablation}, random-budget visual updates already outperform full-image modeling across all metrics, indicating that visual updates yield a more efficient visual representation under a constrained token budget.
We further compare two variants of the TSIM Router, denoted Value and Slope.
Both variants estimate token demand from the empirical relationship among TSIM, token budget, and reconstruction fidelity, but differ in their allocation criteria.
TSIM Router (Value) adopts a value-based criterion, selecting the token budget at which the fitted reconstruction quality reaches a predefined threshold.
In contrast, TSIM Router (Slope) adopts a marginal-utility criterion, selecting the token budget based on the marginal reconstruction gain of additional tokens.

\begin{wraptable}{r}{0.45\linewidth}
\centering

\caption{\textbf{Ablation of \modelname{}'s token allocation strategies on image reconstruction.}}
\label{tab:recon_ablation}

\begin{adjustbox}{max width=\linewidth}
\begin{threeparttable}
\begin{tabular}{l|ccc}
\specialrule{.2em}{.1em}{.1em}

\multicolumn{1}{c|}{Setting}
& rFID$\downarrow$ & PSNR$\uparrow$ & SSIM$\uparrow$ \\ \hline

Full-Image & 1.95 & 28.4 & 0.882 \\
Random Budget & 1.70 & 29.7 & 0.896 \\
TSIM Router (Value) & 1.49 & 29.8 & 0.902 \\
TSIM Router (Slope) & \textbf{1.32} & \textbf{30.4} & \textbf{0.903} \\

\specialrule{.2em}{.1em}{.1em}

\end{tabular}

\begin{tablenotes}[para,flushleft]
\footnotesize
The average number of tokens per image is 64 for all configurations.
``Full-Image'' denotes full-image modeling.
All other settings use visual-update modeling with different token allocation strategies.
`Random Budget’’ allocates token budgets by random sampling from $\mathcal{K}$.
``TSIM Router (Value)'' selects token budgets according to an absolute reconstruction-quality threshold,
whereas ``TSIM Router (Slope)'' selects token budgets according to the marginal reconstruction gain.
\end{tablenotes}

\end{threeparttable}
\end{adjustbox}

\end{wraptable}
As shown in Tab.~\ref{tab:recon_ablation}, TSIM Router (Value) improves over random-budget visual updates, confirming that TSIM provides an effective signal for estimating visual token demand.
However, a value-based threshold considers only the absolute reconstruction quality on the fitted token–fidelity curve and does not explicitly account for the diminishing-return behavior under different TSIM intervals. As a result, the same reconstruction-quality threshold may over-allocate tokens after reconstruction quality has saturated or under-allocate tokens while additional tokens still provide substantial gains.
In contrast, TSIM Router (Slope) directly captures the marginal benefit of additional tokens and selects the budget near the saturation point of the corresponding token–fidelity relationship.
It achieves the best reconstruction performance across all metrics, showing that slope-based allocation provides a slightly more effective TSIM Router criterion than value-based thresholding.

In addition, we ablate the capacity-aware weighting term $K_j$ in TSIM and observe that removing it degrades reconstruction performance, indicating that token budget provides a useful signal for weighting historical visual references. The detailed results are provided in App.~\ref{appendix:k_weight_ablation}.

\begin{figure}[t]
    \centering
    \includegraphics[width=1\linewidth]{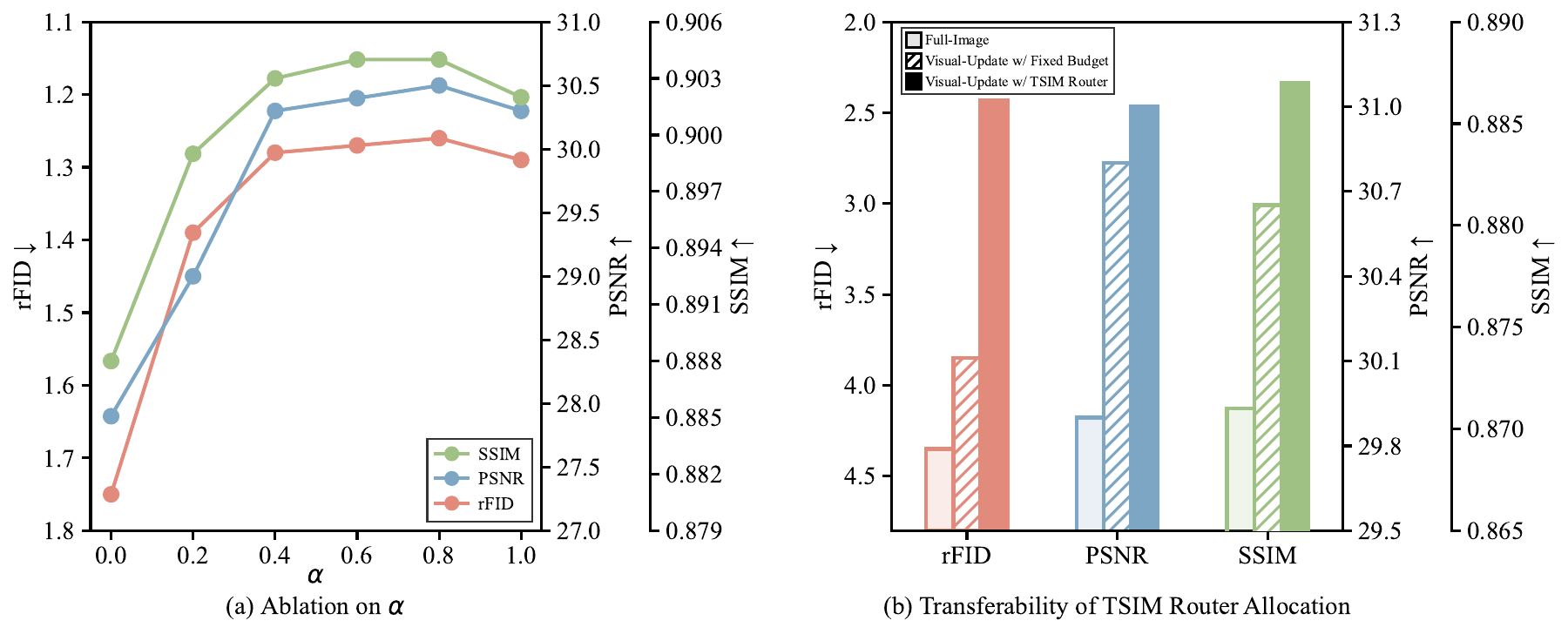}
    \caption{(a) Impact of the time decay coefficient $\alpha$ on TSIM-Router-driven visual updates.
    (b) Transferability of the Zebra-CoT-calibrated TSIM Router allocation rule to StructCoT.}
    \label{fig:combined_ablation}
\end{figure}

\paragraph{Time decay coefficient $\alpha$ ablation.} We conduct an ablation study on the decay parameter $\alpha$ in Eq.~\ref{tsim}, as shown in Fig.~\ref{fig:combined_ablation}(a). Reconstruction performance first improves and then slightly decreases as $\alpha$ increases, indicating a trade-off between preserving long-range historical context and emphasizing temporally local information. When $\alpha=0$, the model relies only on the most recent historical observation, leading to the poorest reconstruction performance. When $\alpha=1$, visual updates from all previous reasoning steps are weighted equally, which overlooks temporal locality and also degrades performance. Intermediate values of $\alpha$, particularly around $0.6$–$0.8$, achieve the best overall performance. This suggests that temporal decay provides a more effective weighting strategy by retaining useful long-range visual context while emphasizing recent visual updates.

\paragraph{Transferability of TSIM-based dynamic allocation.}
To verify whether the empirical mapping between TSIM and visual update token budget observed on Zebra-CoT generalizes to broader scenarios, we apply this mapping to pre-allocate token budgets for StructCoT training samples and train the model accordingly. The metrics on the StructCoT test set are shown in Fig.~\ref{fig:combined_ablation}(b).
Fixed-budget visual updates consistently outperform full-image modeling, suggesting that reusing historical visual representations is also beneficial on StructCoT.
TSIM-Router-driven visual updates achieve the best reconstruction performance across all metrics, reducing rFID from 3.85 to 2.43 compared with fixed-budget visual updates while also improving PSNR and SSIM.
These results indicate that the Zebra-CoT-calibrated TSIM-based allocation remains effective on StructCoT, demonstrating the transferability of the learned TSIM-to-token mapping.

\subsection{Qualitative Comparison}
    \begin{figure*}[t!]
        \centering
        \includegraphics[width=1\linewidth]{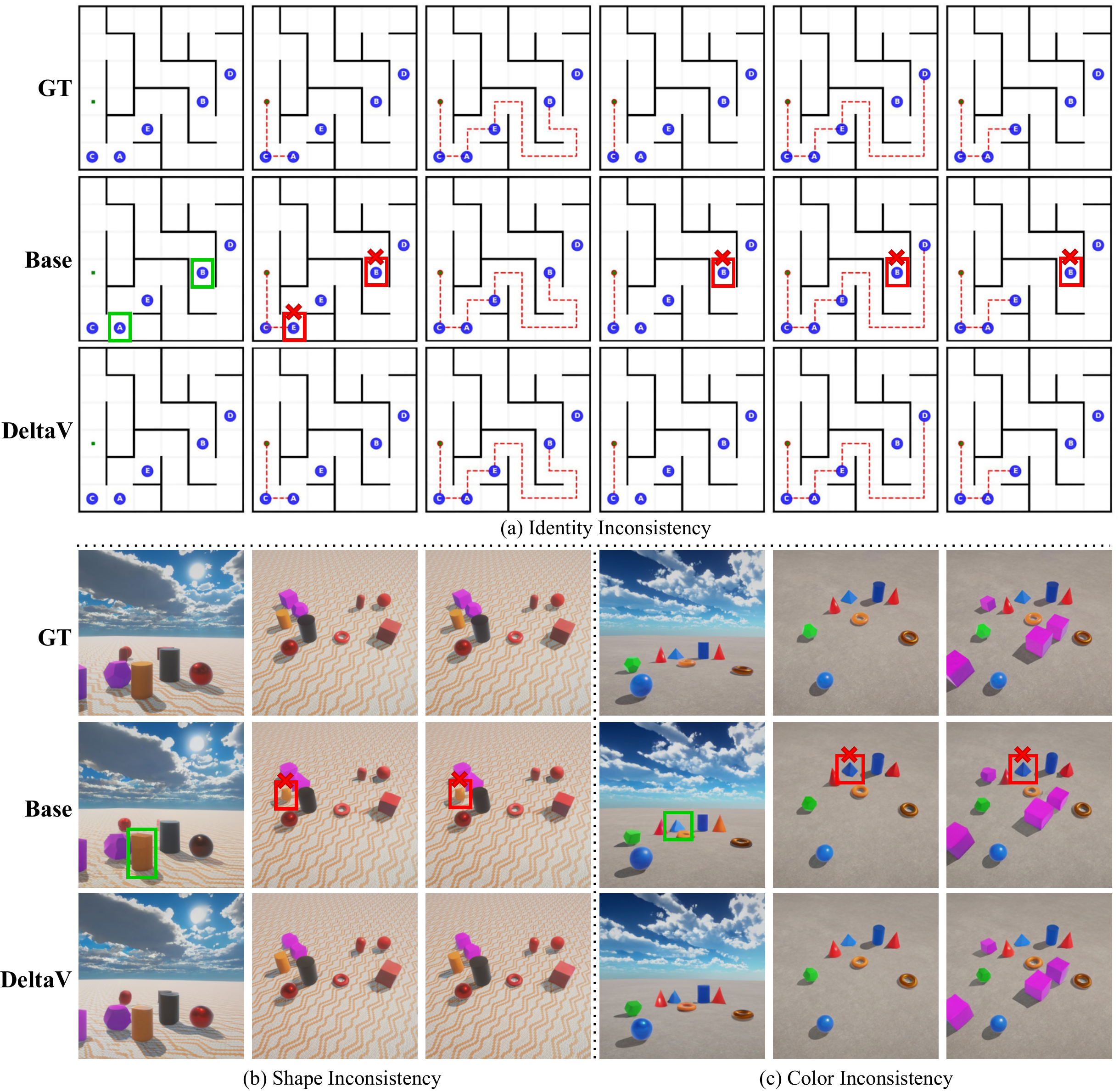}
        \caption{Qualitative comparison of reconstruction results. Full-image modeling (Base) suffers from temporal inconsistencies and semantic drift, including (a) identity inconsistency, where the letter annotation is incorrectly generated; (b) shape inconsistency, where the cylinder is distorted into an irregular object; and (c) color inconsistency, where light blue drifts to dark blue. In contrast, \modelname{} maintains consistent and stable visual representations across reasoning steps.}
        \label{fig:recon_example}
    \end{figure*}
\subsubsection{Qualitative Examples of Image Reconstruction}

We qualitatively compare image reconstruction results between full-image modeling and \modelname{}. As illustrated in Fig.~\ref{fig:recon_example}, full-image modeling exhibits noticeable temporal inconsistencies across reasoning steps. While key visual attributes are correctly reconstructed at earlier steps, they are often not preserved
in subsequent steps,
leading to semantic drift such as identity errors in (a), shape changes in (b), and color changes in (c). Specifically, in case (a), the letter-labeled coordinates are reconstructed correctly in the first image, but A and B are incorrectly replaced by E in the next image, and the error persists throughout subsequent
steps. These observations suggest that full-image reconstruction at each reasoning step makes it difficult to maintain consistency across visual updates.

In contrast, \modelname{} incrementally updates visual representations based on previously reconstructed information, rather than reconstructing each full image. This enables the model to preserve stable visual content while focusing on newly introduced variations, resulting in more consistent reconstructions across reasoning steps.

    \begin{figure*}[t!]
        \centering
        \includegraphics[width=1\linewidth]{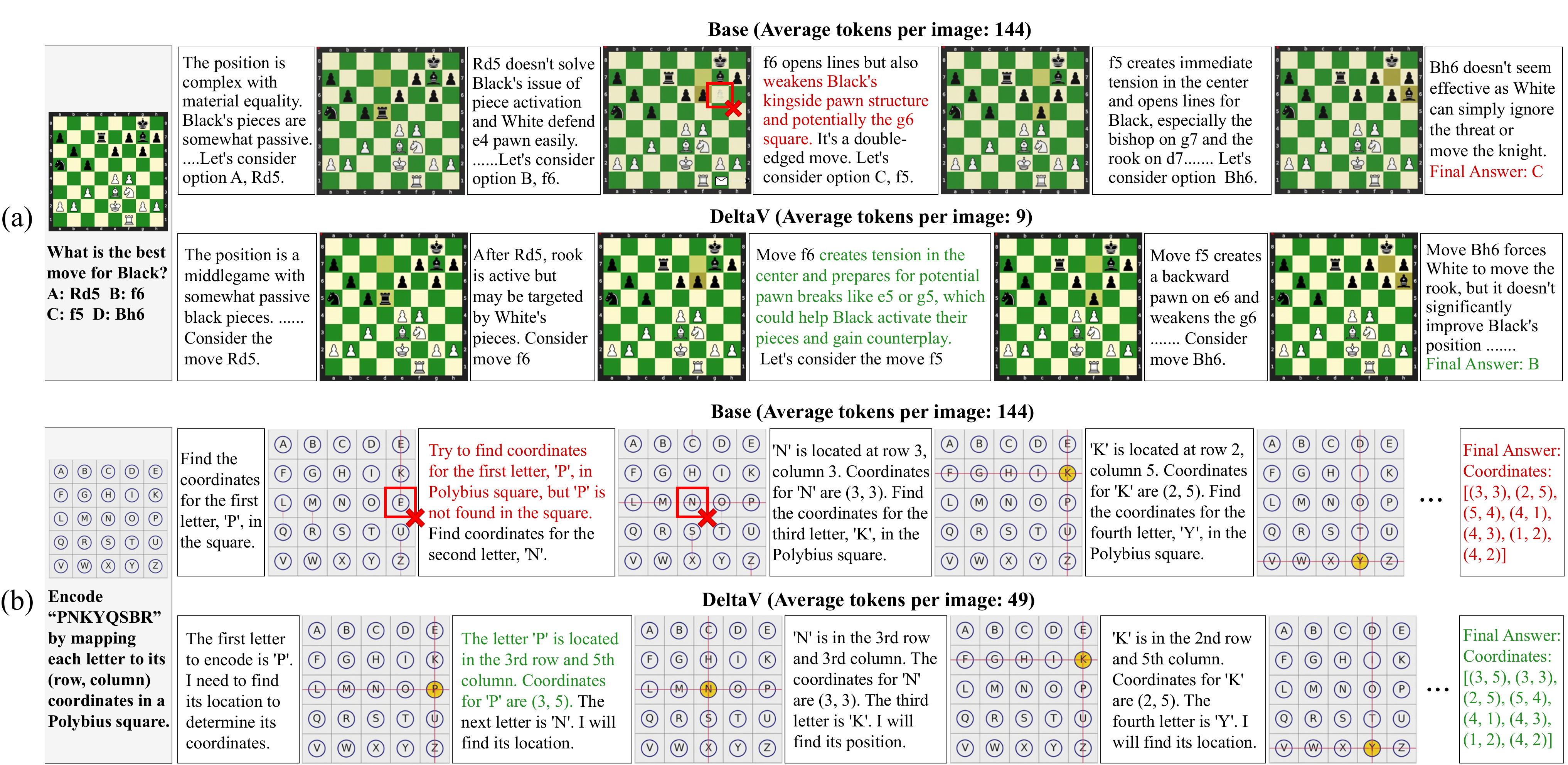}
        \caption{Qualitative comparison of multimodal reasoning. Full-image modeling (Base) exhibits inconsistent intermediate visual states: (a) a missing piece at position g6 leads the model to falsely infer a defensive weakness, and (b) the required letter P is incorrectly reconstructed as E, causing the encoding process to fail. In contrast, \modelname{} maintains consistent visual representations through visual updates.}
        \label{fig:und_example}
    \end{figure*}

\subsubsection{Qualitative Examples of Multimodal Reasoning}

We further examine how different visual modeling paradigms affect multimodal reasoning.
As shown in Fig.~\ref{fig:und_example}(a), full-image modeling omits the piece at position g6, altering the perceived board configuration and leading to an incorrect strategic judgment. Similarly, in Fig.~\ref{fig:und_example}(b), reconstructing the letter P as E in the Polybius square directly causes an incorrect encoding result. These examples show that the effect of visual inconsistency is not limited to reconstruction quality: local errors in intermediate images can alter the semantic evidence used for reasoning and therefore corrupt the subsequent decision process.

In contrast, \modelname{} preserves more consistent visual states by grounding each reconstruction on previously established visual information. This reduces error accumulation across reasoning steps, resulting in more reliable visual evidence and improved reasoning performance.

\subsection{Comparing with Existing Methods}

We compare \modelname{} with representative baselines from four related categories: understanding-centric MLLMs, general ULMMs, latent/perception-token interleaved reasoning methods, and explicit ULMM-based interleaved reasoning methods.
These comparisons cover both in-domain multimodal reasoning and external multimodal reasoning and understanding evaluations.

\begin{table*}[t]
\centering
\caption{\textbf{In-domain multimodal reasoning evaluation. To avoid data leakage, the StructCoT test set excludes all samples originating from Zebra-CoT, and no evaluation samples are included in the training split. All baselines are evaluated under the same input-output protocol and the same answer parsing pipeline.}}
\label{tab:acc_comp}

\begin{adjustbox}{max width=0.98\linewidth}
\begin{threeparttable}

\setlength{\tabcolsep}{4pt}
\renewcommand{\arraystretch}{1.2}

\begin{tabular}{
lc|
*{5}{>{\centering\arraybackslash}m{0.095\linewidth}}|
*{8}{>{\centering\arraybackslash}m{0.095\linewidth}}
}
\specialrule{.2em}{.1em}{.1em}

\multicolumn{2}{c|}{Model} &
\multicolumn{5}{c|}{Zebra-CoT~\citep{li2025zebra}} &
\multicolumn{8}{c}{StructCoT} \\

\multicolumn{1}{c}{Name} & \#Param &
\shortstack{2D} &
\shortstack{3D} &
\shortstack{Science} &
\shortstack{Strategy} &
\shortstack{Overall} &
\shortstack{Strategy\\Planning} &
\shortstack{Spatial\\Planning} &
\shortstack{Logic} &
\shortstack{Math} &
\shortstack{Science} &
\shortstack{Visual\\Search} &
\shortstack{Jigsaw\\Restoration} &
\shortstack{Overall}   \\

\hline

\multicolumn{15}{c}{\textit{Understanding-centric MLLMs}} \\
\hline
GPT-5.2\footnotemark      & -   & 67.6 & 19.3 & 73.3 & 54.4 & 53.7 & 43.1 & 33.8 & 42.1 & 76.3 & 50.4 & 87.0 & 57.1 & 55.7\\
Gemini-3.1-Pro\footnotemark   & -   & 68.7 & 19.0 & 83.3 & 60.4 & 57.9 & 71.6 & 28.2 & 50.2 & 78.3 & 55.0 & 79.4 & 65.3 & 61.1\\
Gemini-3.0-Flash\footnotemark & -   & 66.5 & 19.4 & 78.4 & 54.5 & 54.7 & 55.0 & 33.3 & 44.8 & 74.8 & 48.4 & 83.6 & 64.9 & 57.8\\
Qwen3-VL~\citep{Qwen3-VL}              & 2B  & 44.3 & 13.2 & 30.3 & 9.2  & 24.3 & 3.4  & 31.4 & 4.6  & 41.4 & 29.4 & 80.8 & 39.3 & 32.9 \\
Qwen3-VL~\citep{Qwen3-VL}              & 8B  & 50.7 & 16.9 & \textbf{56.0} & 22.7 & 36.6 & \textbf{21.6} & 25.4 & 13.1 & \textbf{59.3} & 39.3 & 83.8 & 46.5 & 41.3 \\
InternVL3.5~\citep{wang2025internvl3}  & 8B  & 29.7 & 11.4 & 48.9 & 19.8 & 27.5 & 6.9  & 36.3 & 17.5 & 36.1 & 32.0 & 75.8 & 41.0 & 35.1 \\
Qwen2.5-VL~\citep{bai2025qwen25vl}     & 72B & 43.2 & 17.3 & 50.1 & 25.8 & 34.1 & 14.8 & 34.4 & 31.4 & 48.0 & 36.5 & \textbf{84.9} & 47.0 & 42.4 \\

\hline

\multicolumn{15}{c}{\textit{General ULMMs}} \\
\hline
Chameleon~\citep{team2024chameleon}     & 7B  & 13.3 & 3.0 & 5.2  & 9.9  & 7.9  & 5.6 & 12.5 & 4.1  & 9.1  & 13.1 & 23.5 & 14.4 & 11.8 \\
Anole~\citep{chern2024anole}            & 7B  & 10.8 & 2.8 & 4.8  & 8.5  & 6.7  & 5.4 & 0.1  & 3.8  & 8.9  & 12.8 & 16.8 & 11.4 & 9.9 \\
Janus-pro~\citep{chen2025janus}         & 7B  & 31.7 & 7.7 & 11.5 & 18.0 & 17.2 & 4.3 & 24.4 & 13.4 & 16.6 & 12.0 & 74.6 & 33.9 & 25.6 \\
OmniGen2~\citep{wu2025omnigen2}         & 7B  & 26.5 & 1.3 & 9.6  & 9.7  & 11.8 & 0.6 & 25.3 & 1.5  & 8.4  & 10.1 & 78.1 & 28.5 & 21.8 \\
Bagel~\citep{deng2025emerging}          & 7B  & 43.3 & 14.7 & 44.5 & 16.3 & 29.7 & 16.4 & 24.9 & 12.8 & 49.0 & 35.5 & 84.6 & 49.0 & 38.9 \\
EMU3.5~\citep{cui2025emu3}              & 34B & 10.1 & 3.6 & 8.6  & 11.8 & 8.5  & 2.8 & 29.1 & 4.6  & 19.3 & 15.6 & 21.1 & 18.8 & 15.9\\

\hline

\multicolumn{15}{c}{\textit{Latent Interleaved Reasoning Models}} \\
\hline
Monet~\citep{wang2026monet}             & 7B  & 45.4 & 12.5 & 23.2 & 21.9 & 25.8 & 2.1 & 23.3 & 18.9 & 30.5 & 26.3 & 75.4 & 35.8 & 30.3\\
Mirage~\citep{yang2026machine}          & 8B  & 2.9  & 2.6  & 15.5 & 10.8 & 8.0  & 0.9 & 18.1 & 3.0 & 36.8 & 22.6 & 12.0 & 30.4 & 17.7\\
VPT-Det~\citep{yu2025introducing}        & 2B  & 35.7 & 3.6  & 8.1  & 17.9 & 16.3 & 7.5 & 28.3 & 8.8  & 13.9 & 16.5 & 78.9 & 36.0 & 27.1\\

\hline

\multicolumn{15}{c}{\textit{Explicit Interleaved Reasoning ULMMs}} \\
\hline
Bagel-Zebra-CoT~\citep{li2025zebra}      & 7B  & -    & -    & -    & -    & -    & 7.0  & 24.6 & 22.8 & 33.3 & 27.3 & 81.0 & 41.9 & 34.0 \\
ThinkMorph~\citep{gu2025thinkmorph}      & 7B  & 43.0 & 11.6 & 31.4 & 22.9 & 27.2 & 21.4 & 19.5 & 26.4 & 43.4 & 26.0 & 84.1 & 49.9 & 38.7 \\
\rowcolor[HTML]{ECF4FF}
\modelname{}                                      & 2B  & \textbf{78.9} & \textbf{20.0} & 41.1 & \textbf{38.3} & \textbf{44.6} & 16.4 & \textbf{53.0} & \textbf{66.0} & 30.1 & \textbf{45.6} & 84.3 & \textbf{62.6} & \textbf{51.1} \\

\specialrule{.2em}{.1em}{.1em}

\end{tabular}

\begin{tablenotes}[para,flushleft]
\small
\item[2] GPT-5.2: \url{https://chat.openai.com}\quad
\item[3] Gemini-3.1-Pro: \url{https://blog.google/innovation-and-ai/technology/developers-tools/gemini-3-pro-vision/}\quad
\item[4] Gemini-3.0-Flash: \url{https://blog.google/innovation-and-ai/technology/developers-tools/agentic-vision-gemini-3-flash/}\\
\item[] Bagel-Zebra-CoT results on Zebra-CoT are omitted because its training data include the test split.
\end{tablenotes}

\end{threeparttable}
\end{adjustbox}
\end{table*}

\subsubsection{In-domain Multimodal Reasoning Evaluation}

Tab.~\ref{tab:acc_comp} reports in-domain multimodal reasoning results on the Zebra-CoT and StructCoT test sets.
\modelname{}-2B achieves strong overall performance on both datasets, outperforming substantially larger open-source models by 8.0\% on Zebra-CoT and 8.7\% on StructCoT, with an average improvement of 8.4\%.
It also narrows the gap with leading closed-source models and achieves particularly strong results on tasks requiring structured spatial reasoning and progressive visual updates.
Compared with existing models designed for interleaved multimodal reasoning, \modelname{} consistently achieves stronger performance across most task categories. We attribute this improvement to variation-aware visual updates, which reduce redundant modeling of unchanged visual content while preserving visual consistency across reasoning steps.
Note that all baselines are evaluated without fine-tuning on the StructCoT training split; the in-domain comparison therefore reflects the combined effect of the visual-update paradigm and large-scale training.

\subsubsection{External Multimodal Reasoning and Understanding Evaluation}

We further evaluate \modelname{} on a diverse set of external multimodal reasoning and understanding benchmarks, as shown in Tab.~\ref{tab:general_comp}. These benchmarks include reasoning-oriented evaluations such as VStar, EMMA, M3CoT, MathVista, and VisuLogic, as well as understanding-oriented evaluations such as MMBench, MME-P, and MMVP.
Compared with the open-source Qwen3-VL-2B, \modelname{}-2B achieves an average improvement of 5.9\% across these external benchmarks, with MME-P normalized by its 2000-point maximum.
Notably, although \modelname{} is trained primarily for interleaved multimodal reasoning, it also transfers well to external reasoning and understanding benchmarks.
Compared with interleaved-reasoning baselines trained on narrower task distributions, \modelname{} further shows more stable transferability across diverse benchmarks. We attribute this improvement to the combination of StructCoT’s broad coverage of reasoning structures and \modelname{}’s modeling paradigm, which encourages the model to learn reusable visual reasoning patterns rather than task-specific solution strategies.

\begin{table*}[t]
\centering
\caption{\textbf{External multimodal reasoning and understanding evaluation.}}
\label{tab:general_comp}
\renewcommand{\arraystretch}{1.2}

\begin{adjustbox}{max width=0.98\linewidth}
\begin{threeparttable}
\begin{tabular}{
l
c|
*{5}{>{\centering\arraybackslash}m{0.09\linewidth}}|
*{3}{>{\centering\arraybackslash}m{0.09\linewidth}}
}
\specialrule{.2em}{.1em}{.1em}

\multicolumn{1}{c}{\multirow{2}{*}{Model}}
& \multirow{2}{*}{\#Param}
& \multicolumn{5}{c|}{Reasoning}
& \multicolumn{3}{c}{Understanding} \\

\multicolumn{1}{c}{}
&
& VStar & EMMA & M3CoT & MathVista & VisuLogic
& MMBench & MME-P & MMVP \\
\hline

\multicolumn{10}{c}{\textit{General ULMMs}} \\
\hline
Chameleon~\citep{team2024chameleon}         & 7B   & 32.5 & 8.6  & 16.1 & 21.7 & 4.5  & 6.0  & 530  & 4.7  \\
Anole~\citep{chern2024anole}                & 7B   & 34.0 & 6.6  & 15.8 & 22.5 & 3.7  & 6.2  & 508  & 6.7  \\
Janus-pro~\citep{chen2025janus}             & 1B   & 43.5 & 18.9 & 45.9 & 37.6 & 25.0 & 60.2 & 1398 & 39.3 \\
Janus-pro~\citep{chen2025janus}             & 7B   & 39.3 & 21.5 & 49.1 & 42.7 & 17.5 & 66.7 & 1509 & 34.7 \\
OmniGen2~\citep{wu2025omnigen2}             & 7B   & 41.4 & 14.7 & 50.3 & 60.2 & 0.1  & 76.1 & 1588 & 35.3 \\
Bagel~\citep{deng2025emerging}              & 7B   & 70.1 & 28.7 & 31.4 & 72.5 & 28.9 & 83.7 & 1665 & 69.3 \\
EMU3.5~\citep{cui2025emu3}                  & 34B  & -    & -    & -    & 28.3 & 11.4 & 13.7 & 791  & 16.7 \\

\hline

\multicolumn{10}{c}{\textit{Understanding-centric MLLMs}} \\
\hline
Qwen3-VL~\citep{Qwen3-VL}                   & 2B   & 71.7 & 22.2 & 53.0 & 61.1 & 11.5 & 77.1 & 1482 & 45.0 \\
Qwen3-VL~\citep{Qwen3-VL}                   & 8B   & 83.7 & 30.6 & 61.2 & 77.6 & 22.5 & 85.2 & 1729 & 59.3 \\
InternVL3.5~\citep{wang2025internvl3}       & 2B   & 68.1 & 12.7 & 51.3 & 60.8 & 26.0 & 78.2 & 1552 & 48.7 \\
InternVL3.5~\citep{wang2025internvl3}       & 8B   & 69.1 & 16.6 & 59.9 & 74.1 & 29.7 & 82.7 & 1688 & 57.3 \\

\hline

\multicolumn{10}{c}{\textit{Latent Interleaved Reasoning Models}} \\
\hline
Monet~\citep{wang2026monet}                   & 7B  & 79.1 & 22.1 & 44.2 & 62.5 & 10.6 & 75.3 & 1636 & 48.7 \\
Mirage~\citep{yang2026machine}       & 8B   & 13.6 & 13.9 & 1.08 & 29.9 & 0.4 & 12.3 & 549 & 0.0 \\
VPT-Det~\citep{yu2025introducing}       & 2B   & 43.5 & 20.1 & 44.4 & 41.8 & 25.6 & 73.3 & 1516 & 34.0 \\

\hline

\multicolumn{10}{c}{\textit{Explicit Interleaved Reasoning ULMMs}} \\
\hline
Bagel-Zebra-CoT~\citep{li2025zebra}         & 7B   & 64.9 & 20.6 & 62.6 & 72.1 & 0    & 55.6 & 1647 & 22.0 \\
ThinkMorph~\citep{gu2025thinkmorph}         & 7B   & 64.4 & 22.4 & 48.8 & 67.8 & 6.5  & 78.2 & 1478 & 8.6  \\
\rowcolor[HTML]{ECF4FF}
\modelname{}                                           & 2B   & 75.9 & 28.6 & 54.5 & 69.3 & 23.5 & 82.3 & 1555 & 51.3 \\

\specialrule{.2em}{.1em}{.1em}

\end{tabular}
\begin{tablenotes}[para,flushleft]
\footnotesize
Multimodal reasoning benchmarks include
VStar~\citep{wu2024v}, EMMA~\citep{hao2025can}, M3CoT~\citep{chen2024m3cot},
MathVista~\citep{lu2024mathvista}, and VisuLogic~\citep{xu2025visulogic};
Multimodal understanding benchmarks include
MMBench~\citep{liu2024mmbench}, MME-P~\citep{fu2026mme}, and MMVP~\citep{tong2024eyes}.
\end{tablenotes}
\end{threeparttable}
\end{adjustbox}
\end{table*}

\section{Limitation and Discussion}

Our results suggest that visual updates provide an effective paradigm for interleaved multimodal reasoning, where reasoning can be driven by compact state changes rather than by repeatedly modeling full visual states. The proposed token allocation strategy further shows that visual variation can serve as a practical signal for reducing redundant visual modeling. Nevertheless, several limitations remain. First, the resulting budgets should be viewed as an effective empirical allocation strategy rather than as the minimum token requirement across all scenarios; more adaptive token selection algorithms and stronger visual-update modeling architectures remain promising directions for future work. Second, since visual update tokens mainly capture compact changes across visual states, they may underrepresent the fine-grained visual details required for precise perception.
DeltaV currently assumes that the reasoning process can be meaningfully decomposed into intermediate visual states and visual updates. For tasks dominated by pure language reasoning or tasks requiring extremely fine-grained perception, such intermediate visual states may provide limited benefit or may need to be complemented by higher-resolution perception modules or external tools.

In addition, since visual update tokens mainly capture compact changes across visual states, they may underrepresent fine-grained visual details required for precise perception.
This observation points to a broader question: where should the boundary lie between internal representation and external operation in multimodal reasoning? Not all intermediate visual processing needs to be performed through native image generation within the model. For deterministic transformations, such as cropping, resizing, rotation, and local enhancement, external tools can often provide more accurate and cost-effective operations. This suggests that native unified understanding-generation ability and tool-augmented reasoning are complementary rather than mutually exclusive.

Future multimodal reasoning systems should thus move beyond the pursuit of increasingly powerful unified generative capabilities
and learn to adaptively select the appropriate form of intermediate representation.
Depending on the task requirements, a reasoning step may be compressed into a compact visual update, generated as an explicit image, or delegated to an external tool. Such a hybrid framework may offer a better balance among efficiency, controllability, and visual fidelity, while improving generalization to fine-grained perception, precise spatial manipulation, and long-horizon multimodal reasoning.

\section{Conclusion}

In this work, we presented \modelname{}, an update-centric framework for interleaved multimodal reasoning in unified large multimodal models. Instead of treating each intermediate reasoning step as the generation of a complete visual state, \modelname{} formulates visual reasoning as a process of progressive visual state updates. This perspective shifts the learning target from repeatedly reconstructing persistent visual content to modeling the sparse but reasoning-critical changes that drive the evolution of a reasoning trajectory.
Built on this principle, \modelname{} uses compact visual update tokens and a TSIM Router to adapt token allocation to temporal visual variation, reducing redundant visual modeling while preserving reasoning-relevant evidence.
Together with StructCoT, a large-scale dataset covering diverse reasoning structures and visual update patterns, \modelname{} achieves more efficient and effective multimodal reasoning.
Extensive experiments show that it significantly reduces newly generated visual tokens without compromising reconstruction fidelity, improves interleaved multimodal reasoning over full-image generation, and generalizes well to external multimodal reasoning and understanding benchmarks.
More broadly, \modelname{} points to a shift from reasoning over complete visual states to reasoning through visual state transitions, offering a more efficient and generalizable path toward unified multimodal intelligence.

\section*{Acknowledgments}
We sincerely thank Xingchen Liu and Yang Liu for their valuable contributions to the collection, organization, and quality control of the StructCoT dataset.

\bibliography{main}

\clearpage
\appendix
\section{Additional Ablation Studies}
\label{appendix:additional_ablation}
\subsection{Sensitivity to TSIM Router Thresholds}
\label{appendix:tau_sensitivity}

To examine whether the TSIM Router is sensitive to the slope thresholds used in the offline calibration, we evaluate three routed visual-update configurations with different average token budgets. These configurations are instantiated by varying the slope thresholds in the TSIM-to-budget mapping, resulting in average visual-update budgets of 36, 64, and 94 tokens.

As shown in Tab.~\ref{tab:tau_sensitivity}, increasing the average token budget improves reconstruction quality, while the downstream reasoning performance remains relatively stable across different thresholds. In particular, the 64-token setting achieves the best overall reasoning performance, despite using fewer tokens than the 94-token setting. This suggests that allocating more tokens does not necessarily lead to better reasoning, and that removing redundant visual modeling can be beneficial for interleaved multimodal reasoning. The 36-token setting also maintains competitive reasoning performance, but its reconstruction quality is lower. We therefore use the 64-token configuration as the default setting, as it provides a good trade-off between reconstruction fidelity, token efficiency, and reasoning performance.

\begin{table*}[t]
\centering
\caption{\textbf{Sensitivity analysis of TSIM Router thresholds.}}
\label{tab:tau_sensitivity}

\begin{adjustbox}{max width=0.98\linewidth}
\begin{threeparttable}
\begin{tabular}{c|ccc|ccccc}
\specialrule{.2em}{.1em}{.1em}

\multirow{2}{*}{Avg. Tokens}
& \multicolumn{3}{c|}{Reconstruction}
& \multicolumn{5}{c}{Multimodal Reasoning} \\

& rFID$\downarrow$ & PSNR$\uparrow$ & SSIM$\uparrow$
& 2D & 3D & Science & Strategy & Overall \\

\hline

36 & 1.58 & 29.2 & 0.890 & 48.7 & 55.4 & 45.2 & \textbf{43.9} & 48.3 \\
64 & 1.32 & 30.4 & 0.903 & \textbf{48.8} & \textbf{56.0} & \textbf{45.9} & 43.7 & \textbf{48.6} \\
94 & \textbf{1.25} & \textbf{30.9} & \textbf{0.907} & 48.2 & 53.2 & 45.5 & 41.6 & 47.1 \\

\specialrule{.2em}{.1em}{.1em}
\end{tabular}

\begin{tablenotes}[para,flushleft]
\footnotesize
Different average token budgets are obtained by varying the slope thresholds in the TSIM Router calibration. All settings use TSIM-Router-driven visual-update modeling.
\end{tablenotes}

\end{threeparttable}
\end{adjustbox}
\end{table*}

\begin{table}[t]
\centering
\caption{\textbf{Additional ablation studies on TSIM Router design.}}
\label{tab:additional_tsim_ablations}

\begin{subtable}[t]{0.47\linewidth}
\centering
\caption{Amount of TSIM intervals.}
\label{tab:tsim_interval_ablation}
\small
\setlength{\tabcolsep}{4pt}
\renewcommand{\arraystretch}{1.05}
\begin{tabular}{c|ccc}
\toprule
\# $M$ & rFID $\downarrow$ & PSNR $\uparrow$ & SSIM $\uparrow$ \\
\midrule
5  & 1.32 & 30.3 & 0.902 \\
10 & \textbf{1.32} & \textbf{30.4} & \textbf{0.903} \\
15 & 1.33 & 30.2 & \textbf{0.903} \\
\bottomrule
\end{tabular}
\end{subtable}
\hfill
\begin{subtable}[t]{0.50\linewidth}
\centering
\caption{Token-budget weighting term $K_j$.}
\label{tab:k_weight_ablation}
\small
\setlength{\tabcolsep}{4pt}
\renewcommand{\arraystretch}{1.05}
\begin{tabular}{l|ccc}
\toprule
Setting & rFID $\downarrow$ & PSNR $\uparrow$ & SSIM $\uparrow$ \\
\midrule
w/o $K_j$ weighting & 1.38 & 30.1 & 0.900 \\
w/ $K_j$ weighting  & \textbf{1.32} & \textbf{30.4} & \textbf{0.903} \\
\bottomrule
\end{tabular}
\end{subtable}

\end{table}

\subsection{Ablation on the Number of TSIM Intervals}
\label{appendix:interval_ablation}

We study the effect of the number of TSIM intervals used in the offline calibration process.
As shown in Tab.~\ref{tab:tsim_interval_ablation}, using 5, 10, and 15 intervals leads to comparable reconstruction performance, indicating that the TSIM Router is not sensitive to the exact interval amount.
We therefore adopt 10 intervals as the default setting, which provides a stable trade-off between modeling granularity and statistical reliability.

\subsection{Ablation on Capacity-Aware Historical Weighting}
\label{appendix:k_weight_ablation}

In Eq.~\ref{tsim}, the historical similarity term is weighted by both temporal decay and the token budget $K_j$ of each previous visual state.
Here, $K_j$ serves as a lightweight proxy for the preserved visual capacity of state $j$.
To examine whether this capacity-aware weighting is beneficial, we compare it with a variant that removes the $K_j$ term from TSIM:
\begin{equation}
\mathrm{TSIM}^{\mathrm{w/o}\ K}_t =
\frac{
\sum_{j=0}^{t-1} s_{tj} \cdot \alpha^{t-1-j}
}{
\sum_{j=0}^{t-1} \alpha^{t-1-j}
}.
\end{equation}

As shown in Tab.~\ref{tab:k_weight_ablation}, removing the $K_j$ weighting term leads to worse reconstruction performance across all metrics.
This result suggests that capacity-aware weighting helps estimate temporal similarity more reliably.
By emphasizing historical states with richer preserved visual information, the token-budget weighting term leads to more effective visual update token allocation.

\section{Implementation Details of \modelname{}}
\label{app:vimo_implementation_details}

\subsection{TSIM-Tok Tokenizer}
\label{app:visual_update_module}

Given visual states $\{I_0, I_1, \ldots, I_T\}$, we first extract visual features $\{F_0, F_1, \ldots, F_T\}$ using a shared visual backbone. Instead of directly tokenizing the dense feature grid $F_t$, TSIM-Tok uses learnable queries to convert foundation-model visual features into compact discrete slots. We parameterize the query vectors as:
\begin{equation}
V = \{V_0, \Delta V_1, \Delta V_2, \ldots, \Delta V_T\},
\end{equation}
where $V_0 \in \mathbb{R}^{K_0 \times d}$ denotes the learnable queries for the initial visual state, with $K_0$ being the number of base queries and $d$ the query dimension.
Similarly, $\Delta V_t \in \mathbb{R}^{K_t \times d}$ denotes the visual update queries for step $t$, where $K_t$ is dynamically determined by the TSIM Router according to temporal visual variation.
These queries interact with visual features through deformable cross-attention:
\begin{equation}
V_0' = \mathrm{DeformableAttn}
\bigl(Q_{\mathrm{att}} = V_0,\; K_{\mathrm{att}} = V_{\mathrm{att}} = F_0\bigr),
\end{equation}
\begin{equation}
\Delta V_t' = \mathrm{DeformableAttn}
\bigl(Q_{\mathrm{att}} = \Delta V_t,\; K_{\mathrm{att}} = V_{\mathrm{att}} = F_t\bigr),
\quad t=1,\ldots,T.
\end{equation}
The updated queries are concatenated temporally:
\begin{equation}
V' = \{V_0', \Delta V_1', \ldots, \Delta V_T'\}.
\end{equation}
They are further processed by the Difference-aware causal Self-attention Module:
\begin{equation}
\tilde{V} = \mathrm{DiffAwareSelfAttn}\bigl(V'\bigr),
\end{equation}
where $\mathrm{DiffAwareSelfAttn}$ denotes the Difference-aware causal Self-attention Module, and $\tilde{V} = \{\tilde{V}_0,\Delta \tilde{V}_1, \ldots, \Delta \tilde{V}_T\}$.
This difference-aware causal aggregation mechanism is designed to selectively retrieve historical visual information most relevant to the current state, while avoiding redundant modeling of previously encoded visual content.
Consequently, $\Delta \tilde{V}_t$ serves as a compact representation of newly introduced visual information at step $t$, and stable visual structures are implicitly preserved through the tokens from previous steps.

We then quantize both the base visual representation $\tilde{V}_0$ and the visual update representations $\{\Delta \tilde{V}_1, \ldots, \Delta \tilde{V}_t\}$ into discrete tokens:
\begin{equation}
Z_0 = \mathrm{Quantizer}(\tilde{V}_0), \quad
\Delta Z_t = \mathrm{Quantizer}(\Delta \tilde{V}_t), \quad t=1,\ldots,T.
\end{equation}
Let $N_{\mathrm{v}}$ denote the size of the visual codebook.

During TSIM-Tok pretraining, the quantized visual tokens are decoded through a lightweight reconstruction-semantic head. For the visual state at step $t$, the decoder processes the base tokens and all update tokens up to the current step with two groups of learnable decoder queries. The reconstruction queries $V^{\mathrm{rec}}$ and semantic queries $V^{\mathrm{sem}}$ are handled by two readout branches that share the same decoder Transformer:
\begin{equation}
\begin{aligned}
H^{\mathrm{rec}}_t
&= \mathrm{Transformer}_{\mathrm{dec}}
\bigl(Z_0, \Delta Z_{\leq t}, V^{\mathrm{rec}}\bigr),\\
H^{\mathrm{sem}}_t
&= \mathrm{Transformer}_{\mathrm{dec}}
\bigl(Z_0, \Delta Z_{\leq t}, V^{\mathrm{sem}}\bigr),
\end{aligned}
\quad t=1,\ldots,T.
\end{equation}
The reconstruction states $H^{\mathrm{rec}}_t$ are then passed to a lightweight image decoder to obtain the reconstructed image $\hat{I}_t$, while the semantic states $H^{\mathrm{sem}}_t$ are projected to semantic features $\hat{F}_t$ for feature-level supervision.
The tokenizer is optimized with reconstruction, perceptual, adversarial, quantization, and semantic distillation losses:
\begin{equation}
\mathcal{L}_{\mathrm{tok}}
= \lambda_{\mathrm{rec}}\|I_t-\hat{I}_t\|_2^2
+ \lambda_{\mathrm{p}}\mathcal{L}_{\mathrm{LPIPS}}(I_t,\hat{I}_t)
+ \lambda_{\mathrm{adv}}\mathcal{L}_{\mathrm{adv}}
+ \mathcal{L}_{\mathrm{vq}}
+ \lambda_{\mathrm{sem}}\mathcal{L}_{\mathrm{sem}},
\end{equation}
Here, the reconstruction term supervises pixel-level fidelity, $\mathcal{L}_{\mathrm{LPIPS}}$ preserves perceptual similarity, and $\mathcal{L}_{\mathrm{adv}}$ is the adversarial loss used to improve visual realism. The VQ loss learns a stable discrete codebook, and the semantic distillation term aligns $\hat{F}_t$ with the frozen visual-backbone feature $F_t$. The coefficients $\lambda_{\mathrm{rec}}$, $\lambda_{\mathrm{p}}$, $\lambda_{\mathrm{adv}}$, and $\lambda_{\mathrm{sem}}$ balance these objectives. The semantic distillation objective is implemented as feature-level cosine alignment:
\begin{equation}
\mathcal{L}_{\mathrm{sem}}
= 1 - \cos(\hat{F}_t, F_t).
\end{equation}
Specifically, the VQ objective is
\begin{equation}
\mathcal{L}_{\mathrm{vq}}
= \| \mathrm{sg}[z]-z_q \|_2^2
+ \beta \| z-\mathrm{sg}[z_q] \|_2^2
+ \lambda_{\mathrm{ent}}\mathcal{L}_{\mathrm{ent}}.
\end{equation}
Here $z$ denotes the pre-quantized slot feature, $z_q$ denotes its nearest codebook entry, and $\mathrm{sg}[\cdot]$ is stop-gradient. The first term updates the selected codebook entry toward the encoder output, the second term is the commitment loss that keeps the encoder output close to its assigned entry, and $\mathcal{L}_{\mathrm{ent}}$ encourages broader codebook utilization. The coefficients $\beta$ and $\lambda_{\mathrm{ent}}$ control the commitment and entropy regularization terms, respectively.

After TSIM-Tok is trained, \modelname{} uses only its encoder and quantizer to obtain discrete indices for base and visual update tokens. The decoder is not part of the normal \modelname{} autoregressive training or inference path; it is used only for tokenizer reconstruction training/evaluation or optional visualization.

\subsection{Autoregressive \modelname{} Training and Inference}
\label{app:vimo_training_inference}

\modelname{} trains the language model on interleaved text and TSIM-Tok visual tokens. For each training trajectory, TSIM-Tok is frozen and encodes the initial visual state into $Z_0$ and each later visual state into a routed update sequence $\Delta Z_t$ with length $K_t$. The TSIM-Tok visual vocabulary of size $N_{\mathrm{v}}$ is augmented with one visual end token, yielding an $(N_{\mathrm{v}}+1)$-way visual prediction head. We denote the visual end token \texttt{\textless{}|vision\_end|\textgreater{}} used in the main text as $\langle e_{\mathrm{vis}}\rangle$. This token is appended after each generated visual sequence, so the target visual sequence is $(\Delta Z_t,\langle e_{\mathrm{vis}}\rangle)$ for update steps.

Let $\Omega_{\mathrm{text}}$ and $\Omega_{\mathrm{vis}}$ denote supervised text and visual-token positions in the autoregressive sequence. \modelname{} uses standard next-token cross-entropy for both modalities:
\begin{equation}
\mathcal{L}_{\mathrm{text}}
= -\frac{1}{|\Omega_{\mathrm{text}}|}
\sum_{i\in\Omega_{\mathrm{text}}}
\log p_{\theta}(y_i \mid y_{<i}),
\end{equation}
\begin{equation}
\mathcal{L}_{\mathrm{vis}}
= -\frac{1}{|\Omega_{\mathrm{vis}}|}
\sum_{i\in\Omega_{\mathrm{vis}}}
\log p_{\theta}(z_i \mid y_{<i}),
\end{equation}
where $z_i$ is either a TSIM-Tok codebook index or $\langle e_{\mathrm{vis}}\rangle$. The final \modelname{} objective is
\begin{equation}
\mathcal{L}_{\mathrm{\modelname{}}}
= \mathcal{L}_{\mathrm{text}}
+ \lambda_{\mathrm{vis}}\mathcal{L}_{\mathrm{vis}}.
\end{equation}
The two losses are normalized over their valid supervised positions, and $\lambda_{\mathrm{vis}}$ controls the relative weight of visual-token supervision.

During inference, \modelname{} alternates textual reasoning and visual update generation. After a textual step, the visual head autoregressively predicts visual codebook indices until it emits $\langle e_{\mathrm{vis}}\rangle$. Thus the model learns the update length from TSIM-supervised targets, without computing future TSIM scores or invoking the TSIM-Tok decoder during normal reasoning.

\section{StructCoT Dataset Details}
\label{appendix:dataset}
The detailed data distribution of StructCoT is presented in Tab.~\ref{tab:structcot_dataset_detailed}.
The table provides a comprehensive overview of the dataset composition across different task categories and data sources, illustrating the overall coverage and diversity of the dataset. In addition to the statistical distribution, a concise description of the corresponding data collection and preprocessing procedures is included in the table footnote. This supplementary information summarizes the construction pipeline for different subsets of the dataset.
To improve the reliability of StructCoT, we apply source-specific quality control during data construction. For solver-generated tasks, executable symbolic solvers are used to generate intermediate states and final answers, and samples with invalid transitions or inconsistent solutions are removed. For dataset-based QA synthesis, Qwen2.5-VL-72B is employed to filter out ambiguous questions, inconsistent answers, repetitive reasoning processes, and duplicated samples.  In addition, StructCoT is constructed to be template-disjoint from Zebra-CoT, avoiding template-level overlap with the benchmark used for ablation and evaluation. This process reduces noisy supervision while preserving diverse reasoning structures and visual state-transition patterns. For Gemini-synthesized QA subsets, annotations are generated conditionally from images and ground-truth metadata, whereas all evaluated models only access the image-question input during testing.

We further provide detailed descriptions of the reasoning categories summarized in the main text.

\paragraph{Task Categorization by Reasoning Structure.}
To systematically cover the space of multimodal reasoning, StructCoT departs from conventional task- or application-driven taxonomies. Instead, we categorize tasks based on the evolution patterns of visual intermediate states and the computational characteristics of their reasoning processes. This perspective treats visual reasoning as a structured visual update process, where different tasks correspond to distinct modes of state evolution and information transformation.
As illustrated in Fig.~\ref{fig:StructCoT}, we group tasks into seven categories, each representing a distinct visual-language reasoning paradigm:

\begin{itemize}[itemsep=1pt, topsep=1pt]
\item \textbf{Strategic game and planning.} These tasks involve temporally dependent decision-making, where visual intermediates encode evolving states such as board configurations or object transitions. They require long-horizon consistency and coherent policy execution across multiple steps.

\item \textbf{Spatial and embodied planning.} These tasks rely on intermediate representations such as coordinates, viewpoints, and trajectories. Models must reason over dynamically changing environments and integrate multi-view spatial information.

\item \textbf{Logical and abstract reasoning.} Visual intermediates take the form of abstract structures (e.g., diagrams or relational graphs). Solving these tasks requires step-wise decomposition and rule induction over structured visual inputs.

\item \textbf{Mathematical and algorithmic reasoning.} These tasks tightly couple visual representations with symbolic operations, including geometric constructions and algorithmic flows. Models must perform multi-stage reasoning that bridges visual modeling, symbolic computation, and verification.

\item \textbf{Science and domain reasoning.} Visual states encode domain knowledge, such as physical force diagrams, molecular structures, or medical annotations. Effective reasoning requires integrating prior knowledge with dynamically evolving visual evidence.

\item \textbf{Visual search.} These tasks involve iterative attention over visual regions, including scanning, filtering, and localization. They emphasize adaptive allocation of attention between global context and local details.

\item \textbf{Image restoration and jigsaw.} Visual intermediates focus on structure reconstruction, part assembly, and anomaly detection. Models must iteratively refine hypotheses while maintaining global consistency constraints.
\end{itemize}

Representative examples of each task category are illustrated in Fig.~\ref{fig:struct_example}. For categories with limited data availability, we exhaustively collect and curate high-quality samples to maximize coverage. For more abundant categories, we carefully control data volume to maintain a balanced distribution across reasoning types. This design mitigates dominance by any single reasoning pattern and encourages uniform learning signals.

\begin{table*}[t]
\centering
\caption{\textbf{Detailed Statistics of StructCoT Dataset.}}
\label{tab:structcot_dataset_detailed}

\footnotesize
\setlength{\tabcolsep}{4pt}
\renewcommand{\arraystretch}{1.0}

\begin{tabular}{l l r >{\raggedright\arraybackslash}p{6.2cm}}
\toprule
\textbf{Category} & \textbf{Subcategory} & \textbf{\#Samples} & \textbf{Dataset Sources} \\
\midrule

\multirow{9}{*}{\shortstack[l]{Strategic Game\\\& Planning}}
& Checkers$^{\dagger}$ & 25,000 & \multirow{9}{=}{Zebra-CoT + AlgoPuzzleVQA + Enigmata + FrozenLake(HF)} \\
& Chess & 20,500 & \\
& Connect Four & 2,000 & \\
& Tetris & 10,000 & \\
& Move Box$^{\dagger}$ & 22,000 & \\
& Water Jugs$^{\dagger}$ & 22,200 & \\
& Wood Slide$^{\dagger}$ & 22,200 & \\
& Tic-Tac-Toe$^{\dagger}$ & 4,000 & \\
& Frozen Lake & 22,100 & \\

\midrule

\multirow{11}{*}{\shortstack[l]{Logical \&\\Abstract Reasoning}}
& Ciphers & 6,589 & \multirow{11}{=}{Zebra-CoT + PUZZLES + RAVEN + Sudoku + V-Interaction-400K} \\
& Rectangle Tiling$^{\dagger}$ & 21,000 & \\
& Light Up$^{\dagger}$ & 21,000 & \\
& Cube Rolling$^{\dagger}$ & 21,000 & \\
& RAVEN$^{\dagger}$ & 73,000 & \\
& ARC-AGI & 2,000 & \\
& Sudoku$^{\dagger}$ & 10,000 & \\
& Unequal$^{\dagger}$ & 10,000 & \\
& KenKen$^{\dagger}$ & 10,000 & \\
& Unruly$^{\dagger}$ & 10,000 & \\
& Hybrid Logic & 10,800 & \\

\midrule

\multirow{7}{*}{\shortstack[l]{Spatial \&\\Embodied Planning}}
& Maze & 20,000 & \multirow{7}{=}{Zebra-CoT + BridgeV2 + JANUS-VLN + Mobile-R1 + DriveLMM-o1 + 3D-GRAND} \\
& Multi-Hop Spatial Counting & 10,000 & \\
& Robot Planning$^{\ddagger}$ & 43,000 & \\
& Embodied CoT$^{\ddagger}$ & 67,500 & \\
& GUI Agent & 940 & \\
& Autonomous Driving & 17,000 & \\
& Multi-View Perception$^{\ddagger}$ & 40,000 & \\

\midrule

\multirow{8}{*}{\shortstack[l]{Mathematical \&\\Algorithmic Reasoning}}
& Graph Algorithms & 10,000 & \multirow{8}{=}{Zebra-CoT + MathCanvas + Math-VR + MINT-CoT + V-Interaction-400K} \\
& Competitive Programming & 1,207 & \\
& Plane Geometry & 67,000 & \\
& Analytic Geometry & 35,000 & \\
& Solid Geometry & 35,000 & \\
& Algebra & 21,600 & \\
& Calculus & 3,270 & \\
& Statistics & 1,872 & \\

\midrule

\multirow{6}{*}{\shortstack[l]{Science \&\\Domain Reasoning}}
& Chemistry & 11,150 & \multirow{6}{=}{Zebra-CoT + V-Interaction-400K + S-Chain + GAIC} \\
& Physics & 27,100 & \\
& Biology & 9,602 & \\
& Medicine & 10,783 & \\
& Music & 534 & \\
& Aesthetics$^{\ddagger}$ & 13,094 & \\

\midrule

Visual Search
& Visual Search & 86,280 & Zebra-CoT + ThinkMorph + VGR + CoF \\
\midrule

\multirow{7}{*}{\shortstack[l]{Image Restoration\\\& Jigsaw}}
& Jigsaw Puzzle$^{\ddagger}$ & 21,899 & \multirow{7}{=}{Zebra-CoT + COCO2017 + OCR-VQA + TextVQA + Cambrian-10M + LLaVA-OneVision + Infinity-MM} \\
& Missing Piece Selection$^{\ddagger}$ & 19,934 & \\
& Piece Localization$^{\ddagger}$ & 9,980 & \\
& Connection Verification$^{\ddagger}$ & 9,842 & \\
& Order Restoration$^{\ddagger}$ & 19,864 & \\
& Image Restoration (Weather)$^{\star}$ & 49,964 & \\
& Image Restoration (System)$^{\star}$ & 49,899 & \\

\bottomrule
\end{tabular}

\begin{flushleft}
\footnotesize{
$^{\dagger}$ Algorithm-to-image generation: tasks are constructed by optimizing open-source symbolic solvers and rendering intermediate states into images. \\
$^{\ddagger}$ Dataset-based QA synthesis: images and their associated annotations or answers are sourced from existing datasets, and Gemini is used to synthesize question-answer pairs conditioned on these ground-truth annotations.
\\ $^{\star}$ Noise injection: image restoration data are created by applying controlled noise perturbations to clean images.
\\ Note: Zebra-CoT is only included in the training set part of StructCoT.
}
\end{flushleft}
\end{table*}

\section{\modelname{} Training Details}
\label{Training Details}
The training process is divided into four stages in total, with each stage designed under a distinct training configuration. The detailed settings and hyperparameter configurations corresponding to each stage are presented in Tab.~\ref{tab:training_config}. Unless otherwise specified, all images generated or reconstructed by TSIM-Tok are produced at a resolution of $256 \times 256$. In addition, the sources and composition of all training data are summarized in Tab.~\ref{tab:data_sources}, providing an overview of the datasets involved throughout the entire training pipeline. By organizing the training procedure into multiple stages, the framework maintains a structured progression across different phases of optimization and data utilization. Unless otherwise specified, controlled ablation experiments are conducted under matched computational settings. The main \modelname{} training is conducted on 256 GPUs with 80GB memory each, and the ablation experiments are run on 16 GPUs with 80GB memory each.
Specifically, the SigLIP2-Large visual backbone and the Qwen3-based LLM are initialized from Qwen3-VL-2B. The newly introduced components, including the visual update queries, update aggregation layers, quantization-related modules, visual-token projection MLPs, and generation heads, are initialized separately and trained within our framework.

To further improve data quality and training consistency, we employ Qwen2.5-VL-72B as a filtering model during the data preparation process. Specifically, samples containing ambiguous answers, redundant or repetitive reasoning processes, as well as duplicate entries are removed before training. This filtering procedure helps reduce potential noise and inconsistencies in the training corpus while preserving samples with clearer reasoning structures and more reliable supervision signals.

\begin{table*}[t]
\centering
\caption{\textbf{Training Configurations for TSIM-Tok and \modelname{}.}}
\label{tab:training_config}

\small
\setlength{\tabcolsep}{4pt}
\renewcommand{\arraystretch}{1.15}

\resizebox{\textwidth}{!}{
\begin{tabular}{l l c c c c c c c p{5.2cm}}
\toprule
\textbf{Module} & \textbf{Stage} & \textbf{Batch Size} & \textbf{Learning Rate} & \textbf{Max Length} & \textbf{Disc. Weight} & \textbf{Disc. LR} & \textbf{Disc. Start} & \textbf{Training Steps} & \textbf{Data Composition} \\
\midrule

\multirow{3}{*}{TSIM-Tok}
& Single-Image
& 4096 & 1.6e-3 & -- & 0.1 & 4e-4 & 90k & 130k &
\makecell[l]{
Understanding: 60M \\
Generation: 40M \\
Editing: 7M \\
Reconstruction: 7M \\
Cross-modal Interleaved: 1M
}
\\

& Multi-Image
& 4096 & 1.6e-3 $\rightarrow$ 1e-4 & -- & 0.1 & 4e-4 $\rightarrow$ 1e-5 & 10k & 100k &
\makecell[l]{
Single-image: 109M \\
Dual-image: 35M (7M $\times$ 5) \\
Multi-image: 50M (1M $\times$ 50)
}
\\

\midrule

\multirow{3}{*}{\modelname{}}
& Align
& 1024 & 2e-3 & 8192 & -- & -- & -- & 40k &
\makecell[l]{
Understanding: 25M \\
Interleaved: 10M (1M $\times$ 10)
}
\\

& SFT
& 512 & 4e-5 & 8192 & -- & -- & -- & 70k &
\makecell[l]{
Understanding: 25M \\
Interleaved: 10M (1M $\times$ 10)
}
\\

\bottomrule
\end{tabular}
}
\end{table*}

\begin{table}[t]
\centering
\caption{\textbf{Data Sources Across Different Tasks.}}
\label{tab:data_sources}

\small
\setlength{\tabcolsep}{10pt}
\setlength{\linewidth}{0.8\linewidth}
\renewcommand{\arraystretch}{1.6}

\begin{tabularx}{\linewidth}{
>{\centering\arraybackslash}m{5.5cm}
>{\raggedright\arraybackslash}m{6cm}
}
\toprule
\textbf{Task} & \textbf{Data Sources} \\
\midrule

Image Reconstruction
& ImageNet, Flux2 \\

Image Understanding
& OCR-VQA, TextVQA, LLaVA-OneVision, Infinity-MM, FineVision, Cambrian \\

Image Generation
& BLIP3o-60k, JourneyDB, BLIP3o-Pretrain-Long-Caption, Echo-4o-Image, CC-12M, AnyWord-3M, OmniGen2, Mario \\

Image Editing
& ImgEdit, SEED-Data-Edit, AnyEdit, Echo-4o-Image, OmniGen2 \\

Interleaved Multimodal Reasoning
& StructCoT \\

\bottomrule
\end{tabularx}
\end{table}

\section{Metric Evaluation Pipeline}
\label{appendix:Evaluation_Details}
We use deterministic rule-based matching as the primary evaluation protocol. When a model output follows the predefined answer template, the prediction is directly parsed and compared with the ground-truth answer. This protocol is used by default because it is reproducible, efficient, and independent of any additional model-based judgment.

However, model responses do not always strictly follow the required format. Some outputs contain verbose reasoning, minor formatting deviations, or free-form expressions while still providing a clear final answer. Treating all such cases as incorrect would conflate formatting errors with answer correctness. Therefore, we introduce an auxiliary fallback procedure that is activated only when the rule-based parser fails to extract a valid answer. This fallback is not used to re-evaluate normally parsed outputs and does not replace the rule-based metric.

The fallback procedure consists of two constrained LLM-assisted steps, both implemented with the same fixed evaluator, Qwen2.5-VL-72B-Instruct. The first step performs answer extraction: given the raw model response, the evaluator is instructed to output only the final answer claimed by the model, while discarding reasoning traces, explanations, and irrelevant text. This step does not judge correctness. The second step performs semantic matching: given the question, the ground-truth answer, and the extracted answer, the evaluator checks whether the extracted answer is consistent with the ground truth under strict task-specific criteria. Option letters must match for multiple-choice tasks, numerical values must match for counting or calculation tasks, and required action sequences must preserve the same final outcome for procedural tasks. Minor wording or formatting differences are acceptable only when the underlying answer remains semantically equivalent.

\begin{figure*}[t!]
        \centering
        \includegraphics[width=1\linewidth]{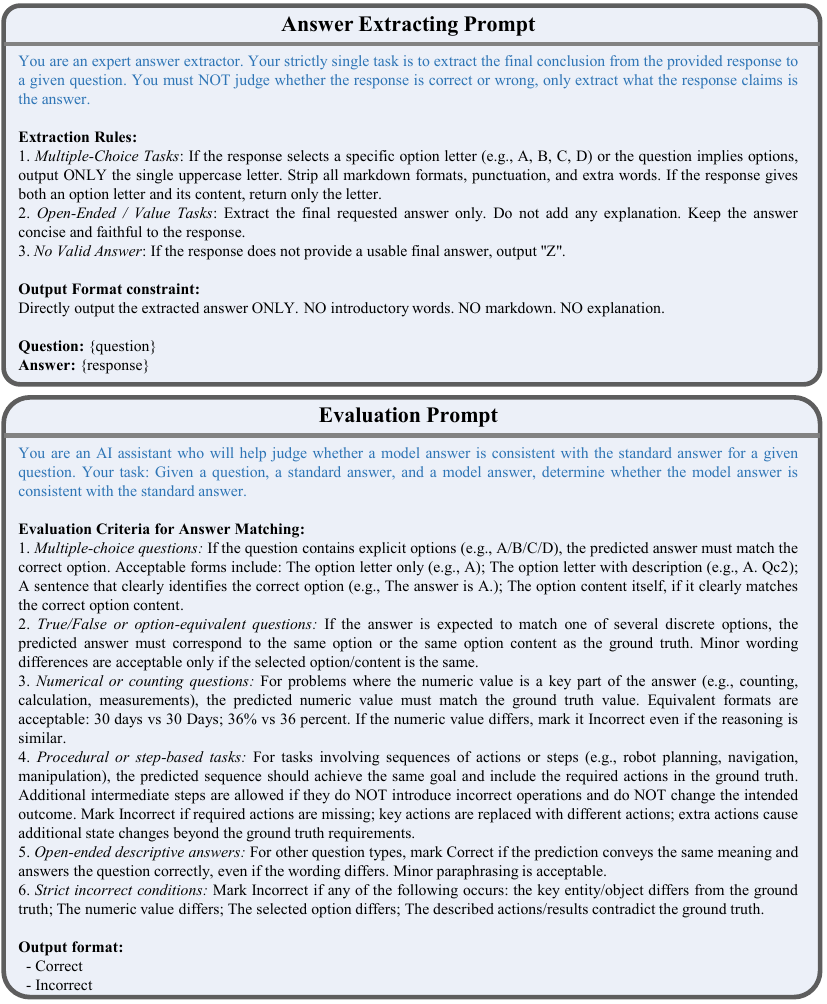}
        \caption{MLLM-based Metric Evaluation Prompts.}
        \label{fig:eval_prompt}
    \end{figure*}

For fairness, the same extraction prompt, evaluation prompt, and evaluator model are applied uniformly across all models and tasks whenever fallback evaluation is required. The evaluator serves only as a standardized verifier for answer extraction and answer matching when rule-based evaluation cannot be applied directly. It does not assess reasoning quality, assign preference scores, rank responses, or use model-specific criteria. Instead, all predictions are evaluated against the same reference answers under a unified evaluation protocol, ensuring that correctness judgments are independent of the model being evaluated. In this way, the evaluation pipeline preserves deterministic rule-based matching for standard outputs while avoiding unfair penalties for responses that are semantically valid but difficult to parse automatically. The detailed prompts are shown in Fig.~\ref{fig:eval_prompt}.

\section{Benchmark Details}
\label{appendix:benchmark_details}

We evaluate our method on eleven representative multimodal reasoning and understanding benchmarks covering fine-grained visual perception, multi-step reasoning, mathematical reasoning, logical reasoning, spatial reasoning, and integrated multimodal understanding abilities. Together, these benchmarks provide a comprehensive evaluation of general multimodal understanding across diverse visual domains and capability dimensions.

\paragraph{EMMA.}
EMMA is an Enhanced MultiModal reasoning benchmark containing 2,788 problems across mathematics, physics, chemistry, and coding, with tasks such as 3D spatial transformations, graph reasoning, reaction simulation, and program visualization. The benchmark is designed to require deep cross-modal reasoning rather than text-only reasoning or a single visual pass. EMMA mainly evaluates fine-grained visual understanding, spatial simulation and integrated visual-textual understanding abilities.

\paragraph{M3CoT.}
M3CoT is a multi-domain, multi-step, multi-modal Chain-of-Thought benchmark designed to test whether models can reason step by step from both visual and textual evidence. It focuses on scenarios where visual information must be actively used during intermediate reasoning rather than only summarized at the input stage. M3CoT mainly evaluates grounded multi-step reasoning, cross-modal evidence integration, and the robustness of multimodal CoT abilities.

\paragraph{VStar.}
VStar is a visual search-oriented benchmark for evaluating whether multimodal models can identify and reason about key details in high-resolution or visually crowded images. It emphasizes guided localization of task-relevant regions before answering visual questions, making it different from standard image-level VQA. VStar mainly evaluates fine-grained perception, visual search, contextual reasoning, and the ability to focus on small but decisive visual evidence.

\paragraph{MathVista.}
MathVista is a mathematical reasoning benchmark in visual contexts, consisting of 6,141 examples collected from 28 existing multimodal datasets and 3 newly created datasets, including geometry diagrams, charts, tables, scientific figures, and puzzle tests. The benchmark organizes evaluation across multiple mathematical reasoning types, tasks, and visual contexts. MathVista mainly evaluates mathematical reasoning abilities in visually intensive scenarios, emphasizing fine-grained visual understanding and compositional reasoning.

\paragraph{VisuLogic.}
VisuLogic is a visual reasoning benchmark with 1,000 human-verified problems across categories such as quantitative shifts, spatial relations, and attribute comparisons. It is designed to reduce language-only shortcuts and place greater emphasis on reasoning directly from visual evidence. VisuLogic mainly evaluates vision-centric logical reasoning, fine-grained comparison, spatial understanding, and compositional visual reasoning abilities.

\paragraph{LogicVista.}
LogicVista is a multimodal logical reasoning benchmark containing 448 multiple-choice questions across five reasoning skills, including inductive, deductive, numerical, spatial, and mechanical reasoning, and nine multimodal capabilities such as diagrams, OCR, graphs, tables, puzzles, and 3D shapes. LogicVista mainly evaluates integrated logical reasoning abilities in visual contexts, emphasizing spatial reasoning, pattern understanding, diagram reasoning, and mechanical reasoning beyond simple recognition tasks.

\paragraph{BLINK.}
BLINK is a benchmark for multimodal large language models that focuses on core visual perception abilities beyond traditional recognition-based VQA. It contains 14 classic computer vision tasks and around 3.8K multiple-choice questions, covering visual correspondence, relative depth estimation, spatial reasoning, multi-view reasoning, jigsaw, visual similarity, and forensics detection. BLINK mainly evaluates nuanced visual perception capabilities and fine-grained visual understanding that cannot be easily reduced to dense captioning or language reasoning alone.

\paragraph{MMBench.}
MMBench is a bilingual multimodal benchmark for objective and fine-grained evaluation of large vision-language models. It uses multiple-choice questions and a circular evaluation strategy to reduce sensitivity to option ordering and answer formatting. MMBench mainly evaluates broad multimodal perception, reasoning, instruction following, and robustness across English and Chinese settings.

\paragraph{MME.}
MME is a comprehensive evaluation benchmark for multimodal large language models, covering both perception and cognition abilities across 14 subtasks. Its instruction-answer pairs are manually designed to support concise and comparable evaluation without heavy prompt engineering. MME mainly evaluates object-level perception, OCR-related understanding, commonsense reasoning, numerical reasoning, and other general multimodal capabilities.

\paragraph{MMVP.}
MMVP is organized around ``CLIP-blind pairs'' — image pairs that are similar in CLIP embedding space but visually different — and constructs 150 image pairs with 300 straightforward VQA questions targeting overlooked visual details. The benchmark summarizes nine basic visual patterns, including orientation, counting, viewpoint, positional relations, and presence of specific features. MMVP mainly evaluates visual grounding ability and models' sensitivity to fine-grained visual patterns and details.

\paragraph{MM-Vet.}
MM-Vet is an evaluation benchmark for complicated multimodal tasks, organized around the integration of six core vision-language capabilities: recognition, OCR, knowledge, language generation, spatial awareness, and math. It defines multiple capability integrations to assess whether large multimodal models can seamlessly combine these abilities to solve open-ended tasks. MM-Vet mainly evaluates integrated multimodal reasoning, complex vision-language understanding, and open-ended generation abilities.

\end{document}